%% file: iclr2023_conference.tex
\documentclass{article} 
\usepackage{iclr2023_conference,times}

\input{math_commands.tex}

\usepackage[colorlinks = true,
            linkcolor = blue,
            urlcolor  = blue,
            citecolor = blue,
            anchorcolor = blue]{hyperref}
\usepackage{url}
\usepackage{graphicx}
\usepackage{color}
\usepackage{subfigure}
\usepackage{wrapfig}
\usepackage{algorithm}
\usepackage{algpseudocode}
\usepackage{authblk}
\floatname{algorithm}{Algorithm}

\newcommand{\tabincell}[2]{\begin{tabular}{@{}#1@{}}#2\end{tabular}}

\title{Prototypical Context-aware Dynamics Generalization for High-dimensional Model-based Reinforcement Learning}

\author[1,2]{Junjie Wang}
\author[3]{Yao Mu}
\author[4]{Dong Li}
\author[1,2]{Qichao Zhang}
\author[1,2]{Dongbin Zhao}
\author[4]{Yuzheng Zhuang}
\author[3]{Ping Luo}
\author[4]{Bin Wang}
\author[4]{Jianye Hao}
\affil[1]{The State Key Laboratory of Management and Control for Complex Systems, Institute of Automation, Chinese Academy of Sciences}
\affil[2]{University of Chinese Academy of Sciences}
\affil[3]{The University of Hong Kong}
\affil[4]{Huawei Noah’s Ark Lab}

\iclrfinalcopy 
\begin{document}

\maketitle

\begin{abstract}
The latent world model provides a promising way to learn policies in a compact latent space for tasks with high-dimensional observations, however, its generalization across { diverse environments with unseen dynamics} remains challenging. Although the recurrent structure utilized in current advances helps to capture local dynamics, modeling only state transitions without an explicit understanding of environmental context limits the generalization ability of the dynamics model. To address this issue, we propose a \textbf{Proto}typical \textbf{C}ontext-\textbf{A}ware \textbf{D}ynamics (\textbf{ProtoCAD}) model, which captures the local dynamics by time consistent latent context and enables dynamics generalization in high-dimensional control tasks.
ProtoCAD extracts useful contextual information with the help of the prototypes clustered over batch and benefits model-based RL in two folds: 1)~It utilizes a temporally consistent prototypical regularizer that encourages the prototype assignments produced for different time parts of the same latent trajectory to be temporally consistent instead of comparing the {features}; 2)~A context representation is designed which combines both the projection embedding of latent states and aggregated prototypes and can significantly improve the dynamics generalization ability. Extensive experiments show that  ProtoCAD surpasses existing methods in terms of dynamics generalization. Compared with the recurrent-based model RSSM, ProtoCAD delivers 13.2\% and 26.7\% better mean and median performance across all dynamics generalization tasks.

\end{abstract}

\section{Introduction}


Latent world models \citep{ha2018recurrent} summarize an agent’s experience from high-dimensional observations to facilitate learning complex behaviors in a compact latent space. Current advances \citep{hafner2019learning, hafner2019dream, deng2022dreamerpro} leverage Recurrent Neural Networks (RNNs) to extract historical information from high-dimensional observations as compact latent representations and enable imagination in the latent space. However, modeling only latent state transitions without an explicit understanding of the environmental context characteristics limits the dynamics generalization ability of the world model. Since the changes in dynamics are not observable and can only be inferred from the observation sequence, for tasks with high-dimensional sensor inputs,  dynamics generalization remains challenging.

Previous works of dynamics generalization on low-dimensional tasks \citep{lee2020context, seo2020trajectory, guo2022relational} offer insight that extracting environmental context information from historical trajectories can benefit both model learning and policy learning, and can improve the generalization ability among different dynamics. However, when dealing with high-dimensional tasks, where the underlying state is not directly accessible, it becomes difficult to apply such methods to extract the environmental context information and rollout the dynamics model for policy planning directly in the observation space. Therefore, in this paper, we investigate how to equip the latent world model with context information about the environment to cope with dynamics generalization.

\begin{figure}[t]
\centering
\subfigure[{RSSM}]{
\begin{minipage}[b]{0.278\textwidth}
\centering
\includegraphics[width=0.95\textwidth]{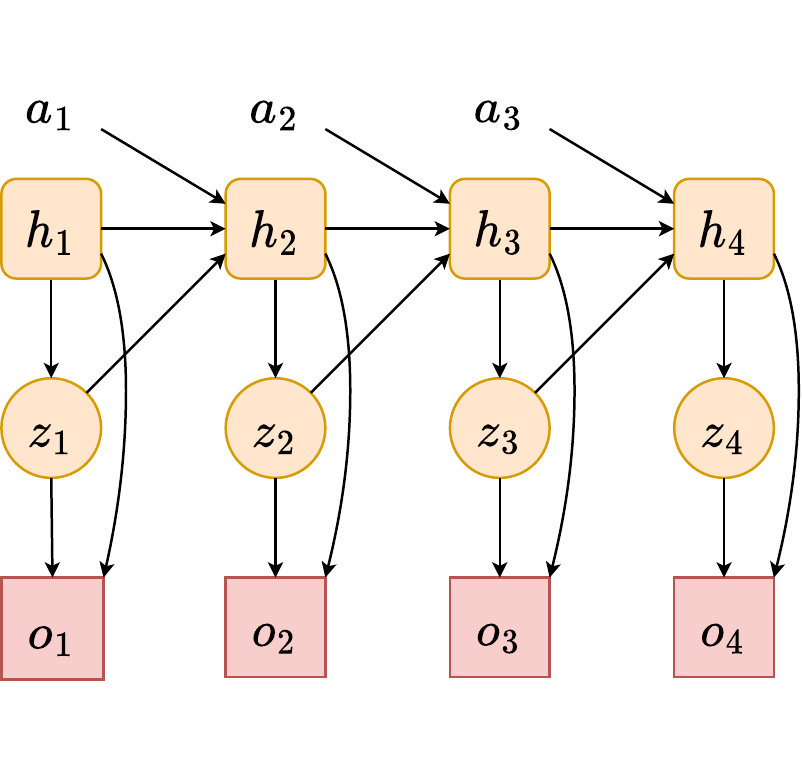}
\end{minipage}
}
\subfigure[ProtoCAD (ours)]{
\begin{minipage}[b]{0.348\textwidth}
\centering
\includegraphics[width=0.95\textwidth]{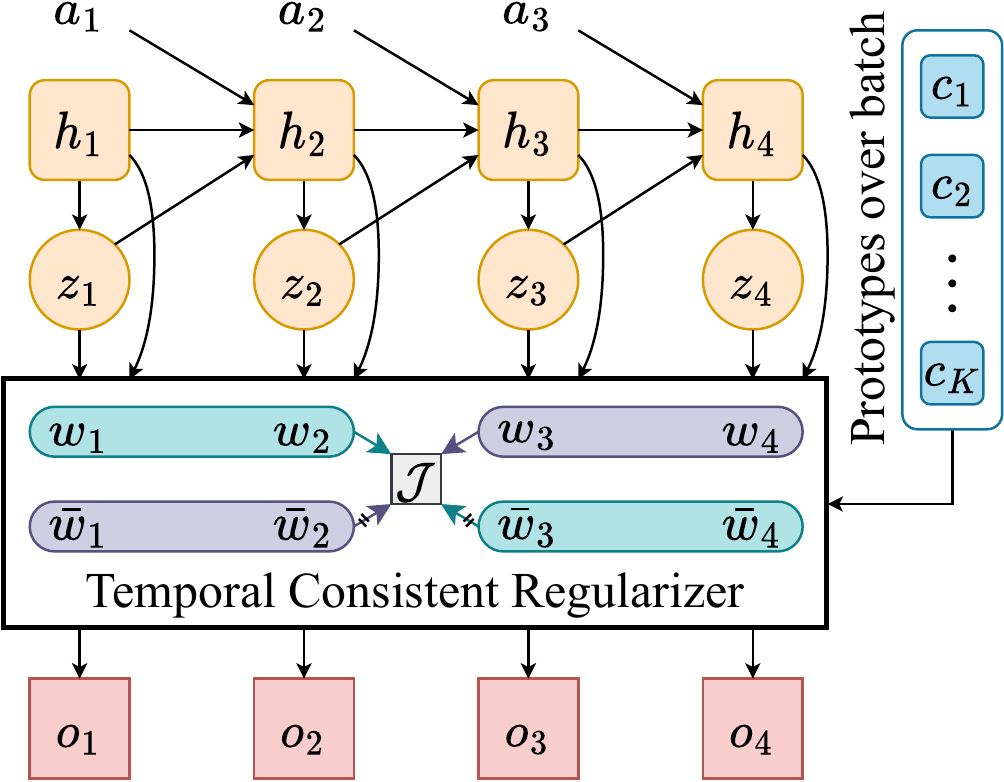}
\end{minipage}
}
\subfigure[{Mean performance at 2M steps}]{
\begin{minipage}[b]{0.32\textwidth}
\centering
\includegraphics[width=1.0\textwidth]{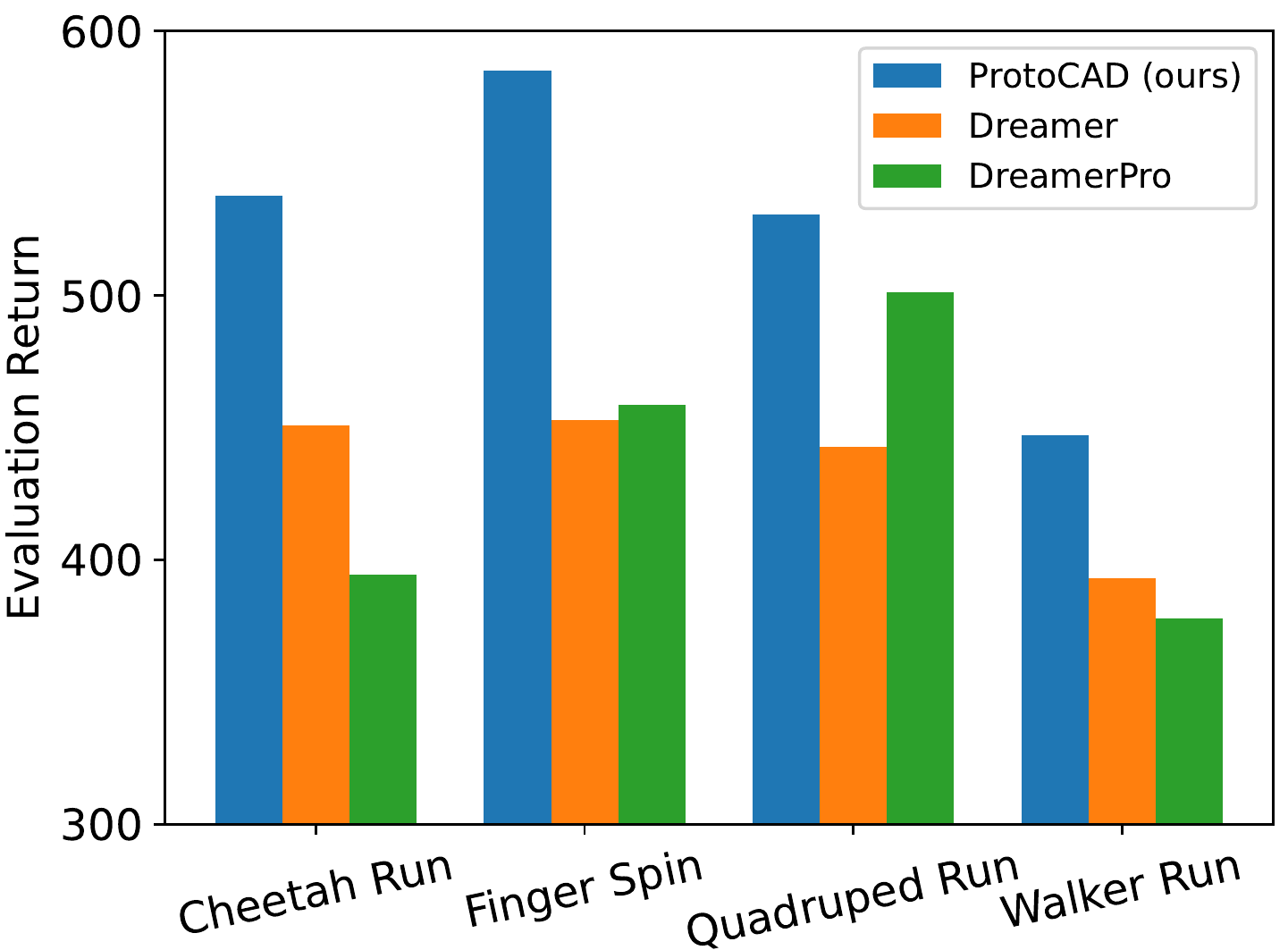}
\end{minipage}
}
\centering
\vspace{-10pt}
\caption{Comparison of different latent world models. We introduce a novel temporally consistent prototypical regularizer into the RSSM~\citep{hafner2019learning} framework. {Here, $h$ and $z$ denote deterministic and stochastic states, respectively, and $o$ denotes observation.} The temporal crossover self-supervised loss provided by this additional regularizer facilitates the extraction of dynamics-related states by the model, while the learned prototypes summarize the characteristics of the seen environments in the experience of the agent. Compared to Dreamer~\citep{hafner2019dream}, which uses RSSM as a dynamics model, and DreamerPro~\citep{deng2022dreamerpro}, which combines RSSM and prototypes for representation, ProtoCAD has better dynamics generalization performance.}
\vspace{-10pt}
\label{intro}
\end{figure}


To reduce the difficulty of extracting contextual information from high-dimensional observations, we present the \textbf{Proto}typical \textbf{C}ontext-\textbf{A}ware \textbf{D}ynamics (\textbf{ProtoCAD}) model, which learns context with the help of prototypes clustered from data. The prototypes summarize the characteristic of dynamics in historical experience and are flexible to extend to unseen dynamics. We enforce temporal consistency between prototype assignments produced for different time parts of the same observation sequence, instead of comparing features directly. Specifically, we first make two different augmentations from the historical observation sequence and embed them into latent
 states. { Then, those latent states are fed into linear projectors to obtain the projection embedding. To extract accurate dynamics-specific information, we regularize the projections and prototypes} to be time consistent in a sequence and invariant to spatial perturbations by a  modified temporal crossover SwAV ~\citep{caron2020unsupervised} loss. By calculating the probability that the { projections} are matched with prototypes, we can obtain an aggregated prototype by combining the learned prototypes with the probabilities as weights. {Finally, both the projection embedding and aggregated prototypes} are combined as the context representation for policy learning to capture contextual information among different dynamics. Figure~\ref{intro} illustrates a brief schematic of ProtoCAD compared to the Recurrent State-Space Model (RSSM), a dynamics model commonly adopted for high-dimensional input tasks.
To the best of our knowledge, this is the first approach that addresses the dynamics generalization problem of high-dimensional inputs. 
The contributions of this work are listed below. 
\begin{itemize}
\item We propose ProtoCAD, a model-based reinforcement learning framework that brings a temporally consistent prototypical regularizer to the latent world model. Benefiting from the backpropagation gradient provided by this additional structure and the designed temporal crossover SwAV loss, the latent model is able to learn more efficient representations for dynamics generalization. 
\item We design a novel context representation that incorporates projection embedding and aggregated prototypes based on the probabilities the projections are matched with prototypes. {The effectiveness of this representation is verified by combining it with latent state as a complete context-based latent feature}. 
\item We develop various visual control environments with different transition dynamics modified from DM-Control Suite to evaluate the performance of ProtoCAD. Extensive experiments demonstrate that our approach achieves better generalization performance on zero-shot visual control tasks. 
\end{itemize}

\section{Related works}

\textbf{Model-based RL with high-dimensional input.}  
Model-based reinforcement learning from high-dimensional observation aims to learn latent representations and policy by latent dynamics models. 
World Model \citep{ha2018recurrent} maps the high-dimensional observations to latent space by VAE ~\citep{pu2016variational} and builds a recurrent latent dynamics model to evolve policy in latent imagination. PlaNet \citep{hafner2019learning} utilizes a recurrent state space world model (RSSM) to learn the representation and latent dynamics jointly. The transition probability is modeled on the latent space instead of the original state space. 
Dreamer~\citep{hafner2019dream} utilizes the RSSM to perform value gradients propagation through long-term imagination. DreamerV2~\citep{hafner2020mastering} extends the Dreamer agent with discrete world model representations. 

\textbf{Self-supervised representation learning.} 
Recent works in self-supervised learning show great potential to learn effective representations from high-dimensional data. One class of these methods learns effective features by comparing positive and negative examples \citep{oord2018representation, chen2020simple,he2020momentum}. MoCo~\citep{he2020momentum} further improves contrastive training by generating the representations from a momentum encoder instead of the trained network. However, these methods need a larger amount of negative samples, which demands large batch sizes or memory banks. To address this challenge, some works propose to learn the representations without discriminating between samples. BYOL~\citep{grill2020bootstrap} introduces a momentum encoder to provide target representations for the training network. SwAV~\citep{caron2020unsupervised} proposes to learn the embeddings by matching them to a set of learned prototypes. 

\textbf{RL with auxiliary visual task.}
Recent works show that self-supervised representation learning techniques are able to improve the performance of visual reinforcement learning significantly. CURL~\citep{laskin2020curl} exacts effective representations via contrastive learning and improves sample-efﬁciency significantly over pixel-based methods. DrQ~\citep{yarats2020image} proposes data-regularized Q-learning, which regularizes the Q-value over multiple image transformations for efficient policy learning. 
Proto-RL \citep{yarats2021reinforcement} conducts a prototypical self-supervised framework that ties representation learning with exploration through prototypes. CTRL \citep{mazoure2021cross} utilizes prototypes to cluster trajectory representations and encourages behavioral similarity between clusters nearby. DreamerPro \citep{deng2022dreamerpro} incorporates prototypes into Dreamer \citep{hafner2019dream} to benefit representation learning and enhance the robustness of reconstruction-free {Model-Based Reinforcement Learning (MBRL)} agents. 
Inspired by these prototypes-based RL methods, we construct a prototypical context learning framework, which extracts temporally consistent contextual information to capture local dynamics efficiently.

\textbf{Dynamics generalization in RL.} 
Dynamics generalization aims to generalize the policy or the learned world model across a distribution of environments with varying transition dynamics. Meta-learning {has} been proposed to improve the generalization ability of RL agents across dynamics changes. Gradient-based meta-RL algorithms \citep{finn2017model, rothfuss2018promp, liu2019taming, gupta2018meta} learn an effective initialization and adapt the policy parameters with few policy gradient updates in new dynamics environments. 
Context-based meta-RL algorithms \citep{rakelly2019efficient, zintgraf2019varibad, lee2020context, seo2020trajectory, fu2021towards, guo2022relational} learn contextual information to capture local dynamics explicitly and show great potential to tackle generalization tasks in complicated environments.
However, the above methods are all investigated in low-dimensional input tasks and lack discussion on the high-dimensional input tasks. We take a step further to build an effective context-based latent dynamics model to solve the dynamics generalization with high-dimensional input.

\section{Problem statement and preliminaries}

\subsection{Problem statement}

Formally, we formulate the problem of high-dimensional control as a discrete-time Partially Observable Markov Decision Process (POMDP), since the underlying state of the environment cannot be obtained directly from the high-dimensional sensory input. A POMDP is a 7-tuple ${\mathcal{M}} \buildrel\textstyle.\over= \left( \mathcal{S}, \mathcal{A}, \mathcal{O}, T, R, P, \gamma \right)$, where $\mathcal{S}$ is the set of states, $\mathcal{A}$ is the set of actions, and $\mathcal{O}$ is the set of observations. 
$T: \mathcal{S} \times \mathcal{A} \to \mathcal{S}$ is the conditional transition probability that action $a_{t} \in  \mathcal{A}$ in state $s_{t} \in \mathcal{S}$ will lead to state $s_{t+1}$. $R: \mathcal{S} \times \mathcal{A} \to \mathbb{R}$ denotes the reward function and $P: \mathcal{S} \times \mathcal{A} \to \mathcal{O}$ denotes the observation probabilities.
\(\gamma \in (0,1)\) is a discount factor. 

Consider a context set $\mathcal{C}$, where different $c \in \mathcal{C}$ {lead to different POMDP models}. For example, the mass of a pendulum can be {considered contextual information}. {Same as in previous works \citep{lee2020context, seo2020trajectory, guo2022relational},} we assume that a context does not change within an episode, only between episodes, and the distribution of contexts is uniform across the context set. For the dynamics generalization problem, the entire set of contexts can be divided into two subsets: $\mathcal{C}_{\text{train}}$ and $\mathcal{C}_{\text{test}}$. In this paper, we focus on zero-shot dynamics generalization, i.e., $\mathcal{C}_{\text{train}} \cap \mathcal{C}_{\text{test}} = \emptyset$. Given a context $c$, we can compute the expected return of policy $\pi$ with $G(\pi ,{{\mathcal{M}}|_c}) \buildrel\textstyle.\over= {\mathbb{E}_\pi }\left( {\sum\nolimits_{t = 0}^\infty  {{\gamma ^t}{r_t}} } \right)$, where ${\mathcal{M}}|_c$ is the POMDP conditioned on $c$. And for any POMDP $M$, we can define the expected return of policy $\pi$ with $\boldsymbol{G}(\pi ,{\mathcal{M}}) \buildrel\textstyle.\over= {\mathbb{E}_{c \sim p(c)}}\left[ {G(\pi ,{{\mathcal{M}}|_c})} \right]$. The goal is to find a policy trained on $\mathcal{C}_{\text{train}}$ that maximizes the expected return on the testing context set: $\mathcal{J}(\pi ) \buildrel\textstyle.\over= \boldsymbol{G}(\pi ,{{\mathcal{M}}|_{\mathcal{C}_{\text{test}}}})$. 

\subsection{Preliminaries}
Many MBRL approaches first learn a world model and then further exploit it to derive policies. Typically, the world model provides a mapping of environmental dynamics from the current state and action to the next state. To extract compact representations from image observation sequences, RSSM \citep{hafner2019learning} separates states into stochastic and deterministic components, allowing the model to robustly learn to predict multiple futures. RSSM is commonly made up of the following components~\citep{hafner2019learning, hafner2020mastering}, 
\begin{align}
\begin{aligned}
&\text{Recurrent module:}&&h_t = f_\phi \left(h_{t-1}, z_{t-1}, a_{t-1}\right)\\
&\text{Representation module:}&&z_t \sim q_\phi \left(z_t \left. {} \right| h_t, o_t \right)\\ 
&\text{Transition module:}&&\hat z_t \sim p_\phi \left(\hat z_t \left. {} \right| h_t \right), 
\end{aligned}
\end{align}
where $o_t$ is the current observation at time step $t$, $h_t$ denotes the deterministic recurrent state, $\hat z_t$, as well as $z_t$, denote the stochastic states of the prior and posterior, respectively, and $\phi$ is the parameter of the model. Here, we denote the deterministic output by $f$, the distribution of samples generated in the real environment by $q$, and their approximation by $p$. The optimization objective of the model is to reduce the KL distance between the prior and the posterior,
\begin{equation}
\mathcal{J}_{\text{RSSM}}^t \buildrel\textstyle.\over= - \beta \text{KL} \left[ q_\phi \left(z_t \left. {} \right| h_t, o_t \right) \left. {} \right\| p_\phi \left(\hat z_t \left. {} \right| h_t \right) \right],
\end{equation}
where $\beta$ is a hyperparameter controlling the loss scale.

\section{Method}

\begin{figure}[t]
\centerline{\includegraphics[width=0.95\textwidth]{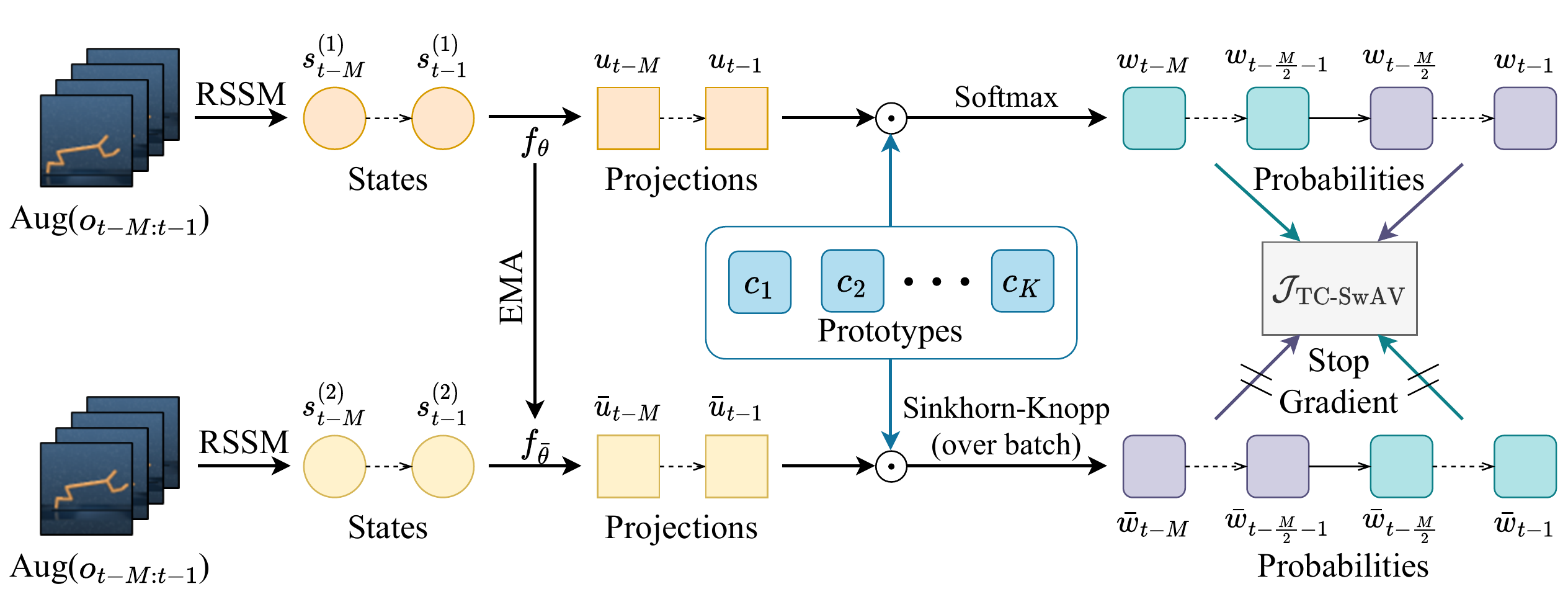}}
\vspace{-10pt}
\caption{
Learning process of prototypical context-aware dynamics model (ProtoCAD). ProtoCAD aims to extract efficient contextual information from high-dimensional observation sequences with the help of prototypes summarized over batch, which summarize the characteristic of dynamics in historical experience. We first make two augmentations from the original observation sequences. Then, the augmented observation sequences are encoded into latent states via RSSM.  Instead of comparing features directly, we enforce temporal consistency between prototype assignments produced for different time parts of the same latent trajectory. Both the projector $f_\theta$ and the prototypes $\{c_k\}_{k = 1}^K$ are updated by the  temporal crossover loss $\mathcal{J}_{\text{TC-SwAV}}$.
}
\vspace{-10pt}
\label{framework}
\end{figure}

In this section, we present the model-based reinforcement learning framework ProtoCAD. 
In order to learn a context-aware world model that facilitates dynamics generalization and policy training, the entire process of ProtoCAD learning is divided into three parts, including latent state encoding, prototypical context learning, and policy optimization. Figure~\ref{framework} provides an overview of the learning process of the prototypical context-aware dynamics model. First, latent state encoding: the raw observations are augmented to obtain two different views, and the two augmented historical trajectories are encoded as latent space states through a transition model RSSM; second, prototypical context learning: linear projections are implemented on the two latent space states to predict cluster assignments, and {we enforce temporal consistency between prototype assignments produced for different time parts of the same latent trajectory. Then we aggregate the prototypes with the probabilities the projections are matched with prototypes as weights.
Finally, the projections and aggregated prototypes are combined as contextual features as a condition for policy optimization.} 
The pseudocode of our overall algorithm is shown in Appendix~\ref{app_algo}. 



\subsection{Latent state encoding}
In this paper, we implement RSSM as the transition model for partially observable environments. 
During the training process, a sequence of historical observations $o_{t-M:t-1}$ and actions $a_{t-M:t-1}$ are sampled from the experience replay buffer, and we first augment the observations to two different views $o^{(1)}_{t-M:t-1}$ and $o^{(2)}_{t-M:t-1}$ with data augmentation. Following DreamerPro \citep{deng2022dreamerpro}, we perform bilinear interpolation of the random shifts and ensure consistency of the augmentation across time steps. Subsequently, augmented observations, together with action sequence are fed into the RSSM to obtain the latent states $s^{(i)}_\tau \buildrel\textstyle.\over= (h^{(i)}_\tau, z^{(i)}_\tau)$, $\tau = t-M, t-M+1, \cdots, t-1$, $i \in \{1,2\}$. 

The deterministic recurrent structure in the transition model combined with stochastic state inference allows it to encode states that both remember multi-step historical information and include the ability to capture environmental uncertainty. This mechanism enables the model to have some generalization capability. However, a simple gradient for latent state updates is insufficient to capture the rich contextual information provided by the environment \citep{lee2020context}, which in turn limits the ability of the model to generalize and transfer policies based on it. Therefore, there is an emerging demand for context-aware dynamics modeling. 

\subsection{Prototypical context learning}

Self-supervised learning approaches show great potential to learn effective representations from high-dimensional data. Among them, SwAV~\citep{caron2020unsupervised} proposes to learn embedding by matching them to a set of learned clusters. Coincidentally, for the generalization problem with a finite number of dynamics, the contextual representations of different environments lie in some clusters. 
In contrast to existing algorithms \citep{yarats2021reinforcement, mazoure2021cross, deng2022dreamerpro} that introduce SwAV into RL to help only represent learning and discard learned prototypes, ProtoCAD groups the latent states into $K$ sets, and combines learned prototypes with projection embeddings to capture contextual information in different dynamics environments. 
We now introduce how to extract prototypical context representations from latent states. 

The $K$ trainable prototypes $\{c_k\}_{k = 1}^K$ can be regarded as corresponding to the $K$ potential ``situations'' in which RL agents may find themselves \citep{mazoure2021cross}. To cluster the latent states output by RSSM into $K$ prototypes, states $s^{(1)}_{t-M:t-1}$ are first fed into a linear projector $f_\theta$ {(we still denote the deterministic output by $f$)} and then $\ell$2 normalized to obtain projections $u_{t-M:t-1}$.  Subsequently, a softmax operation is taken for the dot product of $u_{t-M:t-1}$ and prototypes, 
\begin{equation}
\left( {w_{t-M:t-1,1}, \cdots ,w_{t-M:t-1,K}} \right) = \text{softmax} \left( {\frac{{u_{t-M:t-1} \cdot c_1}}{T }, \cdots ,\frac{{u_{t-M:t-1} \cdot c_K}}{T }} \right).
\end{equation}
where $w_{t-M:t-1,k}$ is the predicted probability that projections $u_{t-M:t-1}$ map to cluster $k$, $T$ is a temperature parameter, and the prototypes $\{c_k\}_{k = 1}^K$ are also $\ell$2 normalized.

To train both the projector and prototypes, we make a copy of the projector, called the target projector~($f_{\bar \theta}$), whose parameters $\bar \theta$ are updated by Exponential Moving Average (EMA) of $\theta$, and cluster its output to compute target projections. 
Since it is ideal to evolve online clustering with the arrival of new batches of trajectories, the Sinkhorn-Knopp~\citep{knight2008sinkhorn} algorithm is adopted, which has been commonly used for online clustering to assist RL tasks~\citep{yarats2021reinforcement, mazoure2021cross, deng2022dreamerpro}. As ProtoCAD is an online operation, Sinkhorn-Knopp is better suited for this task than other clustering methods. Latent states $s^{(2)}_{t-M:t-1}$ are fed into $f_{\bar \theta}$ with $\ell$2 normalization to get target projections $\bar u_{t-M:t-1}$, and target probabilities $\{\bar w_{t-M:t-1,k}\}_{k = 1}^K$ are derived by applying the Sinkhorn-Knopp algorithm to $\bar u_{t-M:t-1}$ and $\{c_k\}_{k = 1}^K$. 

A fundamental characteristic of context is that it does not change within a trajectory. Therefore, we design a temporal crossover SwAV loss to encourage the temporal consistency of the learned features, including projections and prototypes. We divide the above-obtained probabilities $w$ and $\bar w$ into two parts in the time dimension (in the implementation, we set the sequence length $M$ to be even), i.e., 
\begin{equation}
{{y}_{(1),k}} \buildrel\textstyle.\over= {y_{t - M:t - {M \mathord{\left/
 {\vphantom {M 2}} \right.
 \kern-\nulldelimiterspace} 2} -1,k}}, {{y}_{(2),k}} \buildrel\textstyle.\over= {y_{t - {M \mathord{\left/
 {\vphantom {M 2}} \right.
 \kern-\nulldelimiterspace} 2}:t-1,k}}, y \in \{w,\bar w\},
\end{equation}

and then back-propagate the gradient by the following objective, 
\begin{align}
\begin{aligned}
\mathcal{J}_{\text{TC-SwAV}} &\buildrel\textstyle.\over= \frac{1}{2} \sum\limits_{k = 1}^K {\left( {{\bar w_{(1),k}} \cdot \log {w_{(2),k}} + {\bar w_{(2),k}} \cdot \log {w_{(1),k}}} \right)} \\
& = \frac{1}{2} \sum\limits_{k = 1}^K {\left( {\sum\limits_{\tau = t-M}^{t-M/2-1}{\bar w_{\tau,k}}\log {w_{\tau+M/2,k}} + \sum\limits_{\tau = t-M/2}^{t-1}{\bar w_{\tau,k}}\log {w_{\tau-M/2,k}}} \right)}
.
\end{aligned}
\end{align}

Here, TC stands for temporal consistency. Up to this point, we can obtain the 
{aggregated prototypes} 
 of the latent state among the clusters by projector $f_\theta$ and prototypes $\{c_k\}_{k = 1}^K$ with
${e_t} \buildrel\textstyle.\over= \sum\nolimits_{k = 1}^K {{w_{t,k}} \cdot {c_k}}$.  
Then, we obtain context representations including the projection embedding $u_t$ and {aggregated prototypes} $e_t$.
Combined with the latent state $s_t$, this yields a complete latent feature: 
\begin{equation}
x_t \buildrel\textstyle.\over= (s_t, u_t, e_t).
\end{equation}
Considering that the reward function is independent of the environment dynamics (or context), the original observation and reward are predicted from the latent feature space and the latent state space, respectively. That is, 
\begin{align}
\begin{aligned}
&\text{Image predictor:}&&\hat o_t \sim p_\phi \left(\hat o_t \left. {} \right| x_t \right), 
&\text{Reward predictor:}&&\hat r_t \sim p_\phi \left(\hat r_t \left. {} \right| s_t \right). 
\end{aligned}
\end{align}
The distributions produced by the image predictor and reward predictor are trained to maximize the log-likelihood of their corresponding targets, 
\begin{align}
\begin{aligned}
&\mathcal{J}_\text{O}^t \buildrel\textstyle.\over= \ln {p_\phi} ({{\hat o}_t}|{x_t}), &\mathcal{J}_\text{R}^t \buildrel\textstyle.\over= \ln {p_\phi} ({{\hat r}_t}|{s_t}). 
\end{aligned}
\end{align}
To sum up, the overall objective of the prototypical context-aware dynamics model learning is, 
\begin{align}
\begin{aligned}
&\mathcal{J}_\text{ProtoCAD} \buildrel\textstyle.\over= \mathbb{E}_{p_\phi} \left( {\sum\limits_t {\left( {\mathcal{J}_\text{RSSM}^t + \mathcal{J}_\text{O}^t + \mathcal{J}_\text{R}^t} \right)} } + \mathcal{J}_\text{TC-SwAV} \right). 
\end{aligned}
\end{align}

\subsection{Policy learning}


We implement the actor-critic architecture for behavior learning. 
Benefiting from the world model equipped with prototypes, imagination can be performed in the latent space with context features. 
For latent feature $x_t$, the actor and critic models are defined as,
\begin{align}
\begin{aligned}
&\text{Actor:} &&a_t \sim \pi_\psi(a_t|x_t), 
&\text{Critic:} &&v_\xi({x_t}) \approx {\mathbb{E}_{\pi ( \cdot |{x_t})}}\left( {\sum\limits_{\tau  = t}^{t + H} {{\gamma ^{\tau  - t}}{\hat r_\tau }} } \right). 
\end{aligned}
\label{v_cal}
\end{align}
After getting context-aware features of the future through the rollout of the world model, we can further predict the rewards and values of future states and obtain the target values to train actor and critic networks (see more details in Appendix \ref{app policy learning process}). For the estimation of target values, there is a trade-off between model utilization and its prediction accuracy. As the number of model rollout steps increases, the model provides more data for policy training, resulting in higher sample efficiency. At the same time, the accuracy of model prediction decreases. Therefore, we weight the multi-step value estimates to calculate the target value, as in Dreamer, 
\begin{align}
\begin{aligned}
V_N^i({x_\tau}) & \buildrel\textstyle.\over= {\mathbb{E}_{{p_\phi },{\pi _\psi}}}\left( {\sum\limits_{n = \tau }^{h - 1} {{\gamma ^{n - \tau }}{\hat r_n} + } {\gamma ^{h - \tau }}{v_\xi }\left( {{x_h}} \right)} \right) \text{ with } h = \min \left( {\tau  + i,t + H} \right)\\
{V_\lambda }({x_t}) & \buildrel\textstyle.\over= \left( {1 - \lambda } \right)\sum\limits_{n = 1}^{H - 1} {{\lambda ^{n - 1}}V_N^n({x_t}) + } {\lambda ^{H - 1}}V_N^H({x_t}), 
\end{aligned}
\end{align}
where $\tau  = t,t + 1, \cdots ,t + H$.
The learning objectives of the actor and critic models are set as 
\begin{align}
&{\mathcal{J}_\text{Actor}} \buildrel\textstyle.\over= {\mathbb{E}_{{p_\phi},{\pi _\psi}}}\left(\sum\limits_{\tau  = t}^{t + H} {{V_\lambda }({x_\tau })} \right), 
&{\mathcal{J}_\text{Critic}} \buildrel\textstyle.\over= -{\mathbb{E}_{{p_\phi},{\pi _\psi}}}\left( \sum\limits_{\tau  = t}^{t + H} {\frac{1}{2}{{\left\| {{v_\psi }({x_\tau }) - {V_\lambda }({x_\tau })} \right\|}^2}} \right).
\label{LearningObjV}
\end{align}

\section{Experiments}

\subsection{Setups}
\textbf{Environments.} The experimental setup of previous methods \citep{lee2020context, seo2020trajectory, guo2022relational} on the dynamics generalization problem for state inputs provides us with a large number of references. In this paper, we follow many of the environment settings in RIA \citep{guo2022relational}, with the difference that we modify the environment parameters in the DM-Control (DMC)~\citep{tassa2020dmcontrol} benchmark with image-based observations, rather than the standard MuJoCo engine~\citep{todorov2012mujoco} with state observations. Specifically, we conduct our experiments on 8 different visual control tasks, including 3 DMC-Easy benchmark environments (i.e., Hopper Stand, Pendulum Swingup, and Walker Walk) and 5 DMC-Medium benchmark environments (i.e., Cheetah Run, Finger Spin, Hopper Hop, Quadruped Run, and Walker Run). Among these environments, we modify the mass or damping of the components in them and divide all parameter settings into a training set and a testing set. For training and testing, we sample the environment parameters at the beginning of each episode and keep them constant throughout that episode interaction. The parameter settings during testing are not included in the training parameter set. As an example, the experiments in RIA vary the dynamics of HalfCheetah by modifying its rigid link mass and joint damping. Similarly, we achieve different dynamics in the Cheetah Run using the same training and testing parameters (the Cheetah environment in DMC versus the Halfcheetah in MuJoCo). In addition, we also supplement several environments (e.g., Finger and Walker) by referring to existing settings. The specific environmental parameter settings can be found in Appendix~\ref{app_setting}. 

\textbf{Baselines.} To verify the superiority of the proposed ProtoCAD, we compare it with several state-of-the-art~(SOTA) methods, including model-based approaches, as well as model-free methods. \\
(1) Dreamer: Dreamer~\citep{hafner2019dream}  is the state-of-the-art model-based method for high-dimensional control tasks, which constructs a latent world model and proceeds to incorporate multi-step value estimation through imagination in the latent space. Our work builds on Dreamer by introducing prototypes and temporally consistent SwAV loss to capture environmental context information. \\
(2) DreamerPro: DreamerPro~\citep{deng2022dreamerpro} also combines prototypes with Dreamer to extract better representation from the observations, distinct from our approach, they do not extract temporally consistent contextual information to capture local dynamics. \\
(3) Dreamer+Context: 
Since methods like CaDM~\citep{lee2020context} and TMCL~\citep{seo2020trajectory} are difficult to deploy in high-dimensional input tasks, we also implement a context-aware model-based method for a fair comparison, which extracts context  to capture local dynamics by incorporating Dreamer with SwAV~\citep{caron2020unsupervised} and temporal consistency loss. \\
(4) DrQ: DrQ~\citep{yarats2020image} is a SOTA model-free algorithm that applies data augmentation techniques to compute regularized Q values. 

\begin{figure}[t]
\centerline{\includegraphics[width=1.0\textwidth]{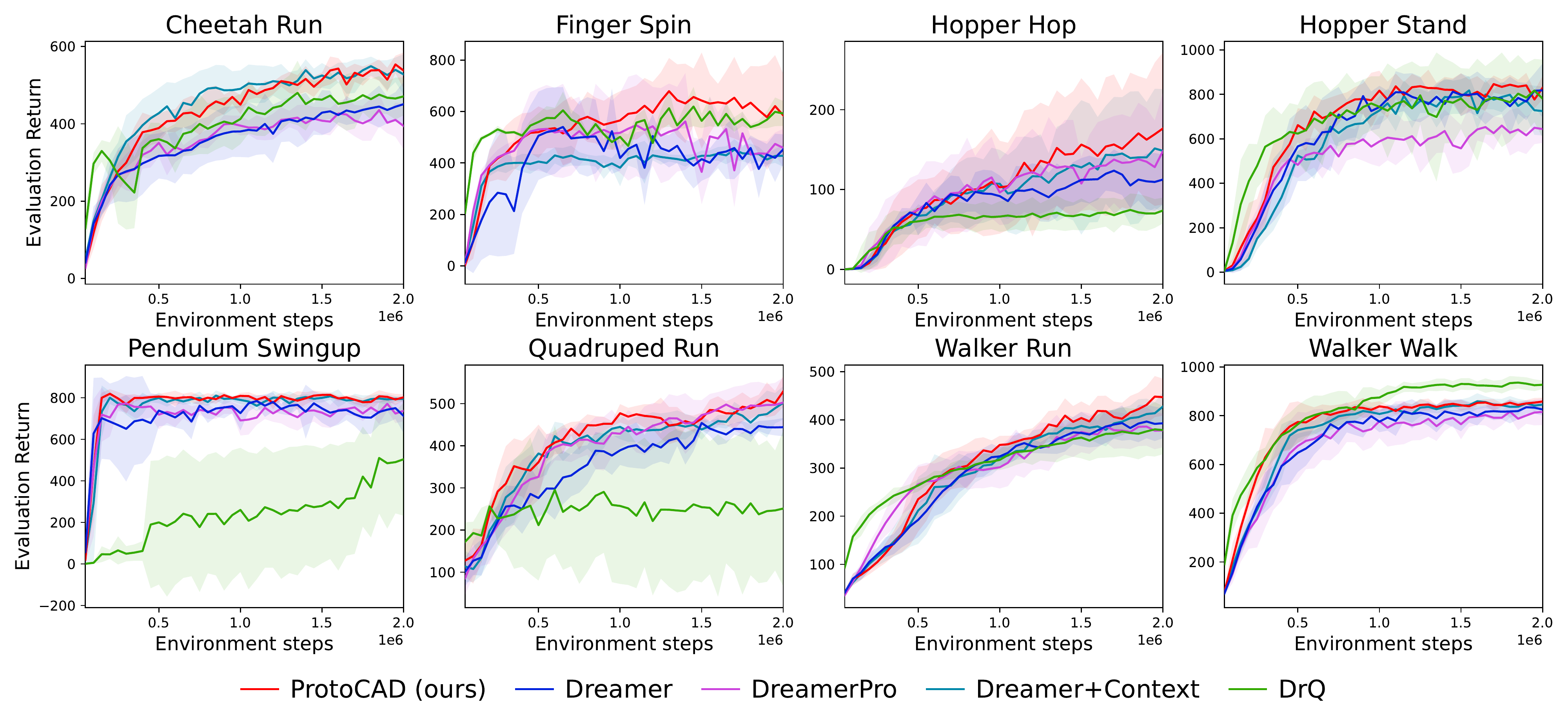}}
\vspace{-10pt}
\caption{Performance comparison of ProtoCAD (ours) and baselines on different zero-shot dynamics generalization continuous visual control environments. The evaluation is performed in environments with unseen dynamics. Solid lines represent the mean score and shaded areas mark the standard deviation across 5 seeds. 
ProtoCAD is comparable to or better than baselines in all tasks. }
\label{main_curves}
\end{figure}

\begin{figure}[t]
\begin{minipage}[b]{0.49\linewidth}
\centering
\includegraphics[width=1.0\textwidth]{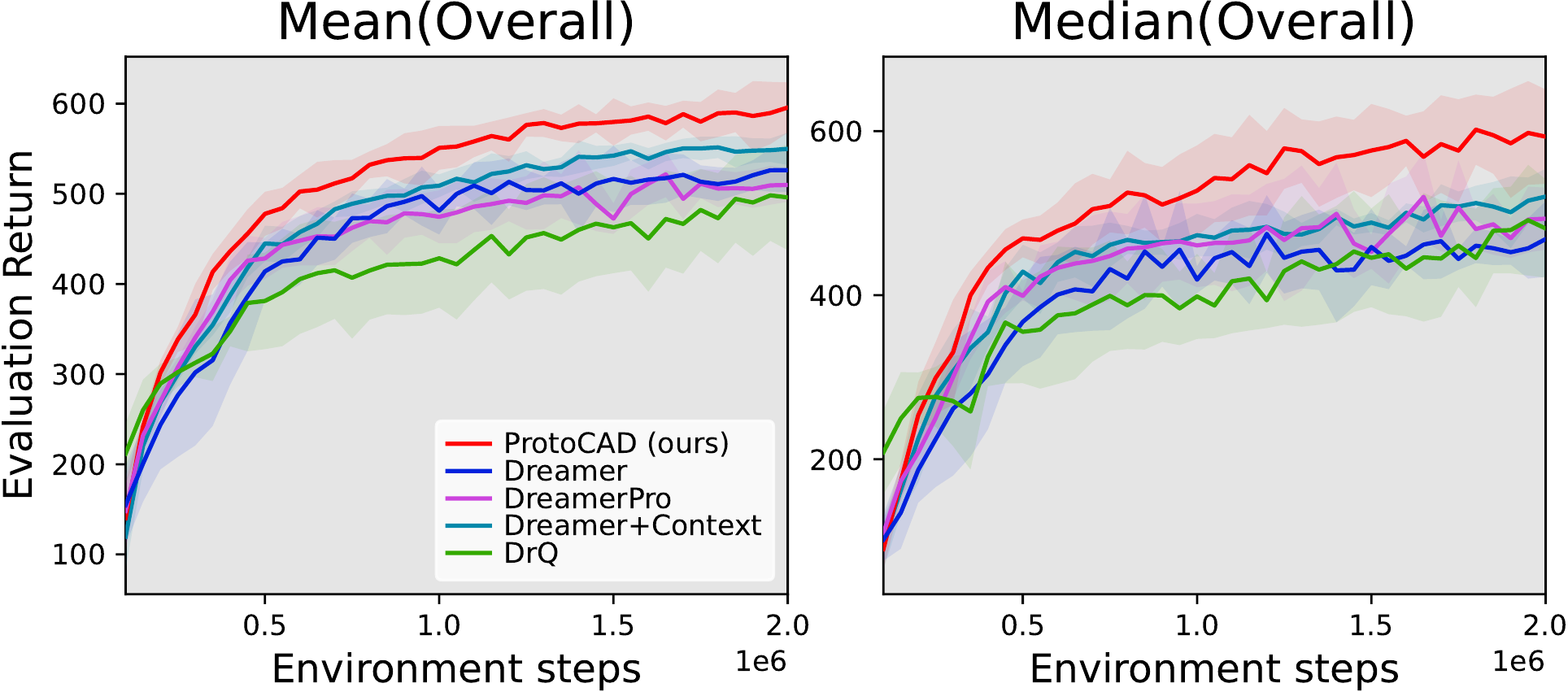}
\vspace{-10pt}
\caption{Overall performance of all methods.}
\vspace{-10pt}
\label{mean_performance}
\end{minipage}
\begin{minipage}[b]{0.49\linewidth}
\centering
\includegraphics[width=1.0\textwidth]{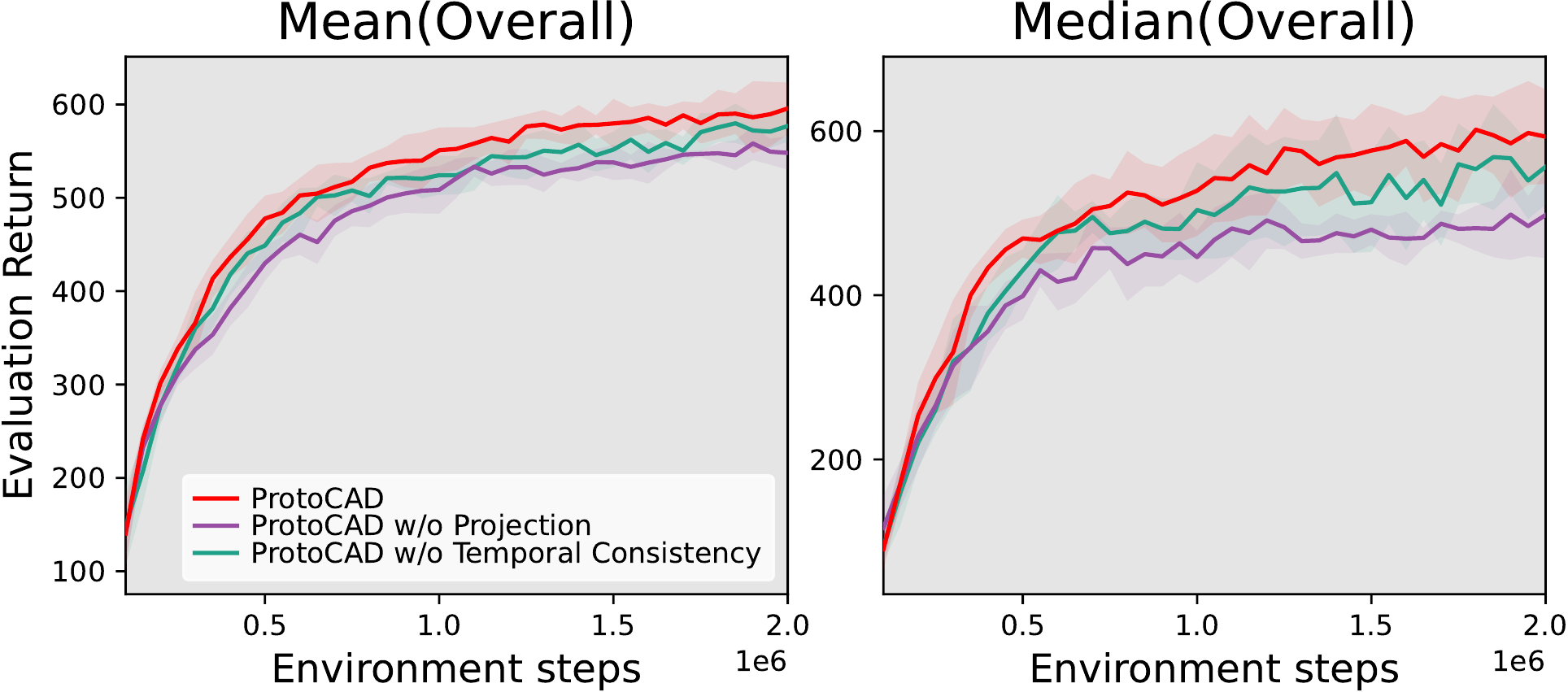}
\vspace{-10pt}
\caption{Overall performance of ablations.}
\vspace{-10pt}
\label{mean_performance_ablation}
\end{minipage}
\centering
\end{figure}

\textbf{Evaluation protocol.} For each task, we train each model in an environment of 2M steps (equivalent to 1M steps for the actor, since the action repeat is set to 2). In test environments (with unseen dynamics), evaluation returns are calculated every 10K steps and averaged over 5 episodes. More training details are given in Appendix~\ref{app_hyper}. 

\subsection{Comparative evaluation}

\begin{figure}[ht]
\centerline{\includegraphics[width=0.95\textwidth]{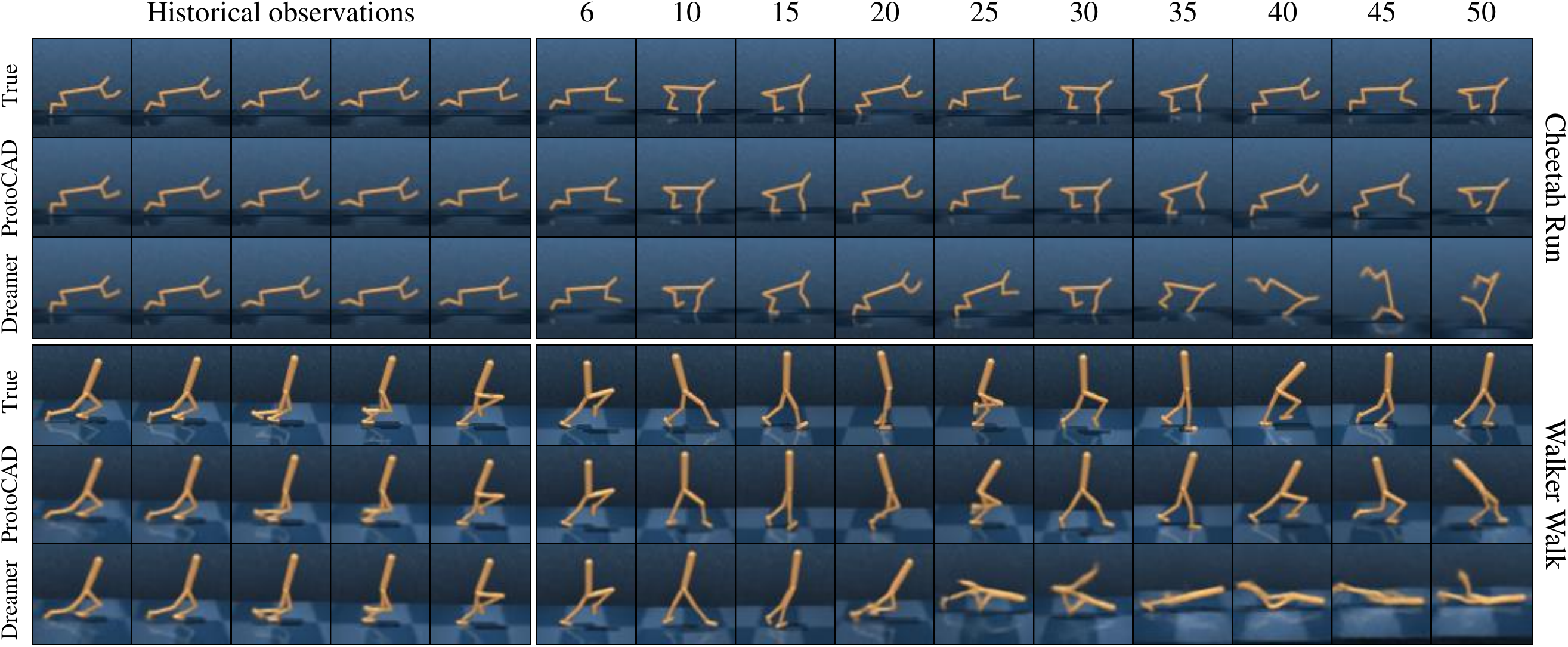}}
\vspace{-5pt}
\caption{Qualitative comparison of long-term predictive reconstructions between ProtoCAD and Dreamer. After the models are trained, we feed them with the same 5 frames of trajectories from the unseen test environment and compare their predicted reconstructions for the next 50 frames with  the true observations. The long-term predictions of ProtoCAD are clearly superior to those of Dreamer. }
\vspace{-10pt}
\label{visual}
\end{figure}

Across diverse visual control experimental tasks, agents trained on the training parameter set by different methods are evaluated under unseen dynamics settings. The performance comparison results are illustrated in Figure~\ref{main_curves} and Table~\ref{compare}. For DreamerPro, we maintain the parameter settings of the original paper with its publicly available code. In all experimental tasks, ProtoCAD exceeds all model-based baselines in terms of sample efficiency and final performance, with significant improvements in Cheetah Run, Finger Spin, Hopper Hop, Pendulum Swingup, Quadruped Run, and Walker Walk. These experimental results suggest that combining prototypes into the latent world model can substantially improve the model's ability to generalize to different transition dynamics. Although DrQ achieves better or comparable performance compared to model-based methods among several methods in Walker Walk and Finger Spin, its performance on Hopper Hop, Pendulum Swingup, and Quadruped Run is noticeably worse than model-based algorithms. Figure~\ref{mean_performance} demonstrates the overall performance of different methods on several experimental tasks, from which it can also be seen that the comprehensive performance of DrQ is inferior to the other methods. The mean and median performance of ProtoCAD on several experimental tasks is significantly better than the other methods. Specifically, compared to Dreamer, ProtoCAD improves the mean and median performance by 13.2\% and 26.7\%, respectively. In addition, Figure~\ref{visual} visualizes the predictive results of ProtoCAD compared to Dreamer in unseen testing environments. It is evident that when facing unseen situations, the predictions of our model have better accuracy, especially for long-term results.

\begin{table}[htb]
  \renewcommand\arraystretch{1.1}
  \centering
  \footnotesize
  \vspace{-10pt}
  \caption{Performance comparison on test environments with unseen dynamics at 2M steps. {Here we use $\pm$ to denote the standard deviation.}}
  \label{compare}
  \vspace{10pt}
\begin{tabular}{l|c|c|c|c|c}
\hline
                 & ProtoCAD (ours)            & Dreamer                    & DreamerPro       & \tabincell{c}{Dreamer+ \\ Context}             & DrQ                       \\
\hline
Cheetah Run      & $\boldsymbol{537.9}_{\pm46.7}$  & $450.9_{\pm57.5}$           & $394.3_{\pm60.2}$ & $528.0_{\pm52.4}$ & $471.1_{\pm49.0}$          \\
Finger Spin      & ${585.0}_{\pm162.1}$ & $453.0_{\pm58.3}$ & $458.5_{\pm74.0}$  & $428.0_{\pm43.0}$ &$\boldsymbol{592.3}_{\pm48.4}$ \\
Hopper Hop       & $\boldsymbol{176.6}_{\pm95.2}$  & $112.6_{\pm31.2}$           & $148.2_{\pm73.1}$ & $148.6_{\pm77.5}$ & $73.7_{\pm16.2}$           \\
Hopper Stand     & $\boldsymbol{829.6}_{\pm53.0}$  & $818.1_{\pm121.1}$ & $644.2_{\pm62.4}$ & $725.2_{\pm110.4}$ & $781.0_{\pm175.4}$          \\
Pendulum Swingup & $\boldsymbol{802.0}_{\pm15.3}$  & $712.8_{\pm82.7}$           & $737.0_{\pm49.9}$  & $794.0_{\pm14.0}$ & $504.0_{\pm272.1}$         \\
Quadruped Run    & $\boldsymbol{529.0}_{\pm36.5}$  & $444.1_{\pm21.7}$           & $502.6_{\pm59.3}$ & $501.9_{\pm39.7}$ & $251.0_{\pm184.6}$         \\
Walker Run       & $\boldsymbol{447.1}_{\pm38.4}$   & $392.9_{\pm31.1}$           & $378.0_{\pm34.9}$ & $427.6_{\pm30.7}$ & $378.7_{\pm50.1}$          \\
Walker Walk      & $858.5_{\pm15.4}$           & $825.6_{\pm24.4}$           & $815.5_{\pm53.6}$ & $845.5_{\pm12.8}$ & $\boldsymbol{927.1}_{\pm17.7}$ \\
\hline
\end{tabular}
\end{table}

To further validate the effectiveness of ProtoCAD, we compare it to DreamerV2, the SOTA model-based approach on the standard DMC benchmark (without varying the dynamics). As shown in Figure \ref{standard_curves} (see Appendix ~\ref{standard}), ProtoCAD also outperforms DreamerV2 in standard DMC benchmarks.

\subsection{Ablation study}
ProtoCAD integrates learned prototypes by predicted probabilities, together with the projection embedding  as a context representation, while using the cross-correspondence between predicted and target probabilities in time sequence to make the context features consistent over a trajectory. We investigate the contribution of each part of ProtoCAD by removing the individual component from it. 

\textbf{ProtoCAD without projection embedding (w/o Projection):} In this setting, the context representation is derived from the combination of prototypes and the latent state, i.e., $x_t = (s_t, e_t)$. 


\textbf{ProtoCAD without temporal dimensional cross-loss (w/o Temporal Consistency):} In this setting, the SwAV loss is calculated from a one-to-one correspondence between the predicted probability $w$ and the target probability $\bar w$ in time sequence, i.e., replace $\mathcal{J}_{\text{TC-SwAV}}$ with $\mathcal{J}_{\text{SwAV}} = \frac{1}{2} \sum\limits_{k = 1}^K {\left( {{\bar w_{(1),k}} \cdot \log {w_{(1),k}} + {\bar w_{(2),k}} \cdot \log {w_{(2),k}}} \right)} $. 


We plot the mean and median performance for all tasks in Figure~\ref{mean_performance_ablation}. Please refer to Appendix~\ref{app_ablation} for curves of all individual tasks. It can be seen that each composition contributes significantly to the performance of ProtoCAD, combining all of them achieves the best results across different tasks. 

To further investigate the adaptability of prototypes to dynamics generalization problems, we implement an additional context-aware model-based agent by combining Dreamer with BYOL~\citep{grill2020bootstrap}. The results are given in Appendix~\ref{app_byol}, from which it can be seen that the extraction of context features from the latent space by self-supervision alone is not effective enough. This is further evidence that the prototypical approach is applicable to the dynamics generalization problem. 

\section{Conclusion}
Dynamics generalization with high-dimensional observations is a critical yet challenging problem in model-based reinforcement learning. In this paper, we propose a novel model-based framework,  \textbf{Proto}typical \textbf{C}ontext-\textbf{A}ware \textbf{D}ynamics (\textbf{ProtoCAD}) model, which introduces prototypes into the latent world model while simultaneously performing latent space representation learning and temporally consistent context clustering. Evaluations of challenging visual control tasks with unseen dynamics demonstrate that our approach achieves state-of-the-art performance in terms of sample efficiency and final score. 
Our ablation experiments further illustrate that the superiority of ProtoCAD is attributed to the context representation that combines projections and prototypes as well as the temporal consistency loss. 
In addition, extending our framework by incorporating advanced task-relevant information extraction techniques to further improve the dynamics generalization capability could be left as our future work.

\bibliography{iclr2023_conference}
\bibliographystyle{iclr2023_conference}

\newpage

\appendix

\section{Appendix}

\subsection{Algorithm}
\label{app_algo}

The training pseudocode is given in Algorithm~\ref{algo}.

\begin{algorithm*}[ht]
  \caption{Prototypical Context-aware Dynamics (ProtoCAD)}
  \begin{algorithmic}[1]
    \State Initialization: Number of random seed episodes $N$, collect interval $C$, batch size $B$, sequence length $M$, Number of prototypes $K$, temperature parameter $T$, imagination horizon $H$, episode length $L$, learning rate $\alpha$
    \State Collect dataset $\mathcal{D}$ with $N$ episodes through the interaction with the environment \texttt{ENV} using random actions
    \State Initialize world model parameters $\phi$ and $\theta$, prototypes $\{c_k\}_{k = 1}^K$, actor network parameters $\psi$, critic network parameters $\xi$
    \State Initialize $\bar \theta = \theta$
    \While{not converged}
      \For{$c = 1,\cdots, C$}
        \State Sample $B$ data sequences $\left\{ {\left( {{a_{\tau}},{o_{\tau}},{r_{\tau}}} \right)} \right\}_{\tau = t-M}^{t} \sim \mathcal{D}$
        \State Perform data augmentation to obtain $\left\{ {\left( {{a_{\tau}},{o^{(i)}_{\tau}},{r_{\tau}}} \right)} \right\}_{\tau = t-M}^{t}, i \in \{1,2\}$
        \State Derive RSSM states, $ h^{(i)}_\tau = f_\phi \left(h^{(i)}_{\tau-1}, z^{(i)}_{\tau-1}, a_{\tau-1}\right) $, $z^{(i)}_\tau \sim q_\phi \left(z^{(i)}_\tau \left. {} \right| h^{(i)}_\tau, o^{(i)}_\tau \right)$
        \State Concatenate states $s^{(i)}_\tau = (h^{(i)}_\tau, z^{(i)}_\tau)$
        \State Compute $u_\tau = {f_\theta }({s^{(1)}_\tau}) $ and $\left( {w_{\tau,1}, \cdots ,w_{\tau,K}} \right) = \text{softmax} \left( {\frac{{u_{\tau} \cdot c_1}}{T }, \cdots ,\frac{{u_{\tau} \cdot c_K}}{T }} \right)$
        \State Compute $\bar u_\tau = {f_{\bar \theta} }({s^{(2)}_\tau}) $ and $\left( {\bar w_{\tau,1}, \cdots , \bar w_{\tau,K}} \right) = \text{Sinkhorn-Knopp} \left( {\bar u_{\tau}, \{c_k\}_{k = 1}^K} \right)$
        \State Update $\phi$, $\theta$ and $\{c_k\}_{k = 1}^K$ using $ \mathcal{J}_\text{ProtoCAD} $
        \State Update $\bar \theta$ by exponential moving average of $\theta$
        \State Imagine trajectories $\{(s_\tau, a_\tau)\}^{t+H}_{\tau=t}$, $u_\tau = {f_\theta }({s_\tau}) $, ${e_\tau} = \sum\nolimits_{k = 1}^K {w_{\tau, k} \cdot {c_k}} $
        \State Concatenate features $x_\tau = (s_\tau, u_\tau, e_\tau)$ and compute value estimates $V_\lambda ({x_\tau})$
        \State Update actor network parameters \(\psi  \leftarrow \psi  + \alpha {\hat \nabla _\psi }{\mathcal{J}_\text{Actor}}\)
        \State Update critic network parameters \(\xi  \leftarrow \xi  - \alpha {\hat \nabla _\xi }{\mathcal{J}_\text{Critic}}\)
      \EndFor
      \State ${o_1} \leftarrow \texttt{ENV.reset()}$
      \For{$t = 1,\cdots, L$}
        \State Compute $ h_t = f_\phi \left(h_{t-1}, z_{t-1}, a_{t-1}\right) $, $z_t \sim q_\phi \left(z_t \left. {} \right| h_t, o_t \right)$ from history and $s_t = (h_t, z_t)$
        \State Compute $u_t = {f_\theta }({s_t}) $ and ${e_t} = \sum\nolimits_{k = 1}^K {\text{softmax} \left(\frac{{{u_t} \cdot {c_k}}}{T }\right) \cdot {c_k}} $
        \State Get latent feature $x_t = (s_t, u_t, e_t)$ and get \(a_t \sim \pi_\psi(a_t|x_t)\) with the actor
        \State Add exploration noise to action and execute it to get \({r_t,o_{t+1}} \leftarrow \texttt{ENV.step(}a_t\texttt{)}\)
      \EndFor
      \State Add experience to dataset \(\mathcal{D} \leftarrow \mathcal{D} \cup \left\{ {\left( {{a_t},{o_t},{r_t}} \right)} \right\}_{t = 1}^L\)
    \EndWhile
  \end{algorithmic}
  \label{algo}
\end{algorithm*}

\subsection{The imagination process for policy learning}
\label{app policy learning process}
\begin{figure}[ht]
\centerline{\includegraphics[width=0.9\textwidth]{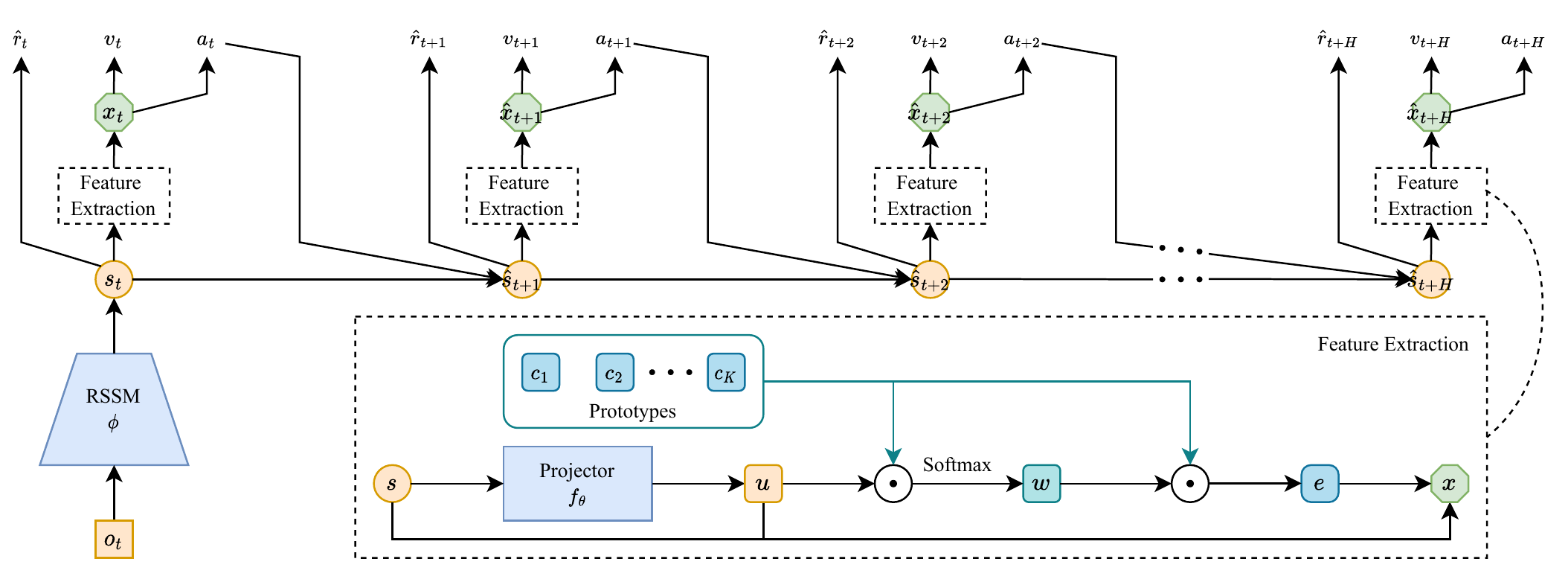}}
\caption{Latent trajectory imagination for policy learning.}
\label{fig:policy learning}
\end{figure}

As shown in Figure \ref{fig:policy learning}, with the prototypical context-aware dynamics model, we can imagine the latent trajectories by model rollout without any interaction with the environment.  We can further predict the rewards and values of future states and obtain the target values based on the context representation and latent states. The policy is optimized under the actor-critic framework.

\subsection{Environmental settings}
\label{app_setting}
We follow the main environmental settings of RIA~\citep{guo2022relational} in dynamics generalization. The difference is that we vary the dynamics in the visual observation of DMC rather than the state observation of MuJoCo in RIA. Details of the setting are given below. 
\begin{itemize}
    \item Cheetah: We refer to the setting of Halfcheetah in RIA to change its dynamics by modifying the mass of rigid link $m$ and the damping of joints $d$. 
    \item Hopper: We refer to the setting of Hopper in RIA to change its dynamics by modifying the mass $m$ of the hopper agent. 
    \item Pendulum: We refer to the setting of Pendulum in RIA to change its dynamics by modifying the  mass $m$ of the pendulum. 
    \item Quadruped: We refer to the setting of Ant in RIA to change its dynamics by modifying the mass of quadruped's leg $m$ to change its dynamics. 
\end{itemize}
In addition, we supplement two environments, Finger and Walker, with references to Hopper and Cheetah settings, respectively. That is,
\begin{itemize}
    \item Finger: We refer to the setting of Hopper to change its dynamics by modifying the mass $m$ of the finger agent. 
    \item Walker: We refer to the setting of Cheetah to change its dynamics by modifying the mass of rigid link $m$ and the damping of joints $d$. 
\end{itemize}

The specific environmental parameter settings are listed in Table~\ref{env_para}.

\begin{table}[htb]
  \renewcommand\arraystretch{1.0}
  \centering
  \caption{The environmental settings.}
  \label{env_para}
\begin{tabular}{l|l|l}
\hline
          & Training Parameter List & Test Parameter List \\
\hline
Cheetah   & \tabincell{c}{$m \in \{0.75,0.85,1.00,1.15,1.25\}$ \\ $d \in \{0.75,0.85,1.00,1.15,1.25\}$} & \tabincell{c}{$m \in \{0.2,0.3,0.4,0.5,$ \\ $\qquad \quad 1.5,1.6,1.7,1.8\}$ \\ $d \in \{0.2,0.3,0.4,0.5,$ \\ $\qquad 1.5,1.6,1.7,1.8\}$} \\
\hline
Finger    & $m \in \{0.5, 0.75, 1.0, 1.25, 1.5\}$                                      & $m \in \{0.25, 0.375, 1.75, 2.0\}$                                                         \\
\hline
Hopper    & $m \in \{0.5, 0.75, 1.0, 1.25, 1.5\}$                                      & $m \in \{0.25, 0.375, 1.75, 2.0\}$                                                         \\
\hline
Pendulum  & \tabincell{c}{$m \in \{0.75,0.8,0.85,0.9,0.95,1.0,$ \\ $\quad 1.05,1.1,1.15,1.2,1.25\}$} & \tabincell{c}{$m \in \{0.2,0.4,0.5,0.7,$ \\ $\qquad \quad 1.3,1.5,1.6,1.8\}$}\\
\hline
Quadruped & $m \in \{0.85,0.90,0.95,1.00\}$                                             & \tabincell{c}{$m \in \{0.20,0.25,0.30,0.35,0.40,$ \\ $\quad 0.45,0.50,0.55,0.60\}$}\\
\hline
Walker    & \tabincell{c}{$m \in \{0.75,0.85,1.00,1.15,1.25\}$ \\ $d \in \{0.75,0.85,1.00,1.15,1.25\}$} & \tabincell{c}{$m \in \{0.2,0.3,0.4,0.5,$ \\ $1.5,1.6,1.7,1.8\}$ \\ $d \in \{0.2,0.3,0.4,0.5,$ \\ $1.5,1.6,1.7,1.8\}$} \\
\hline
\end{tabular}
\end{table}

\subsection{Hyperparameters}
\label{app_hyper}

For hyperparameters that are shared with DreamerPro~\citep{deng2022dreamerpro}, we use the default values suggested in the config file in the official implementation of DreamerPro. With the following two exceptions: we set the batch size as 16 as in Dreamer, and the number of prototypes $K$ to be task-specified. The main hyperparameters are listed in Table~\ref{hyperpara} and Table~\ref{setting_k}. 

\begin{table}[htb]
  \centering
  \caption{Hyperparameters setting. }
  \label{hyperpara}
  \begin{tabular}{c|c|c}
    \hline
    Hyperparameter & Meaning & Value \\ 
    \hline
    $OP$ & Optimizer & Adam \\ 
    $N$ & Number of random seed episodes & $2$ \\
    $C$ & Collect interval & $100$ \\
    $B$ & Batch size & $16$ \\
    $M$ & Sequence length & $50$ \\
    $A$ & Action repeat & $2$ \\
    $\gamma$ & Discount factor & $0.99$ \\
    $\alpha_w$ & Learning rate of the world model & $3\times 10^{-4}$ \\
    $\alpha_a$ & Learning rate of the actor model & $8\times 10^{-5}$ \\
    $\alpha_c$ & Learning rate of the critic model & $8\times 10^{-5}$ \\
    $H$ & Imagination horizon & $15$ \\
    \hline
    $D$ & Prototype dimension & $32$ \\
    $T$ & Softmax temperature & $0.1$ \\
    $SK-itr$ &  Sinkhorn-Knopp iterations & $3$ \\
    $SK-eps$ &  Sinkhorn-Knopp epsilon & $0.05$ \\
    $\eta$ & Momentum update fraction & $0.05$ \\
    \hline
  \end{tabular}
\end{table}

\begin{table}[htb]
  \centering
  \caption{Hyperparameter setting of the number of prototypes $K$. }
  \label{setting_k}
\begin{tabular}{l|c}
\hline
Task             & Value\\
\hline
Cheetah Run      & 100 \\
Finger Spin      & 100 \\
Hopper Hop       & 50  \\
Hopper Stand     & 100 \\
Pendulum Swingup & 100 \\
Quadruped Run    & 100 \\
Walker Run       & 50  \\
Walker Walk      & 100 \\
\hline
\end{tabular}
\end{table}



\subsection{Additional results}


\subsubsection{Standard DMC comparison}
\label{standard}
\begin{figure}[ht]
\centerline{\includegraphics[width=1.0\textwidth]{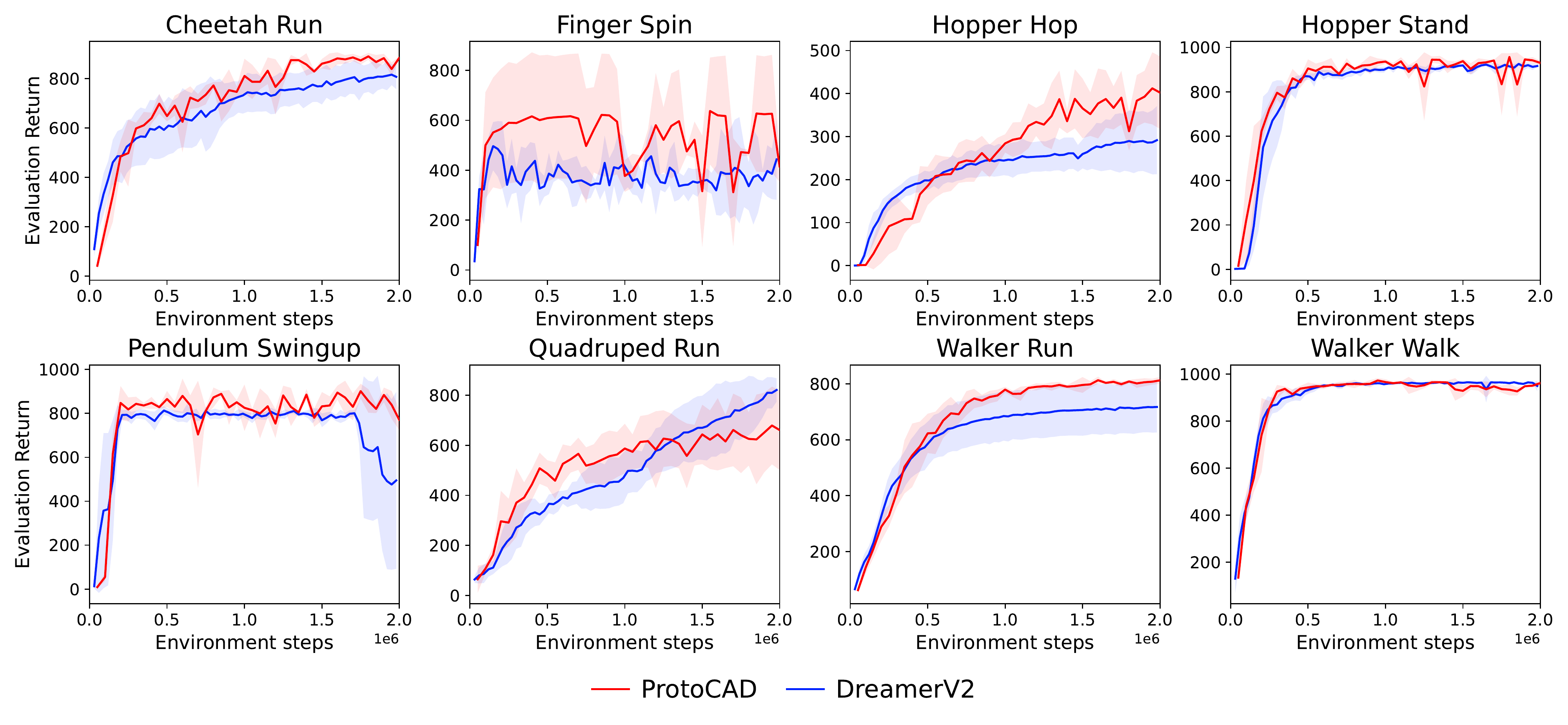}}
\caption{Performance comparison of ProtoCAD and DreamerV2 on standard DMC tasks (without dynamics variation). The mean and standard deviation of ProtoCAD are computed from 3 seeds.}
\label{standard_curves}
\end{figure}

Figure~\ref{standard_curves} shows the performance of our approach versus DreamerV2~\citep{hafner2020mastering} on standard DMC tasks (without dynamics variation). 
For a fair comparison with DreamerV2, here all our model-related parameter settings are kept the same as its open-source code. 
The DreamerV2 results are from the original open-source repository. ProtoCAD outperforms DreamerV2 by a large margin. This also indicates that our approach works for different versions of the world model. 

\subsubsection{Ablation results of ProtoCAD}
\begin{figure}[ht]
\centerline{\includegraphics[width=1.0\textwidth]{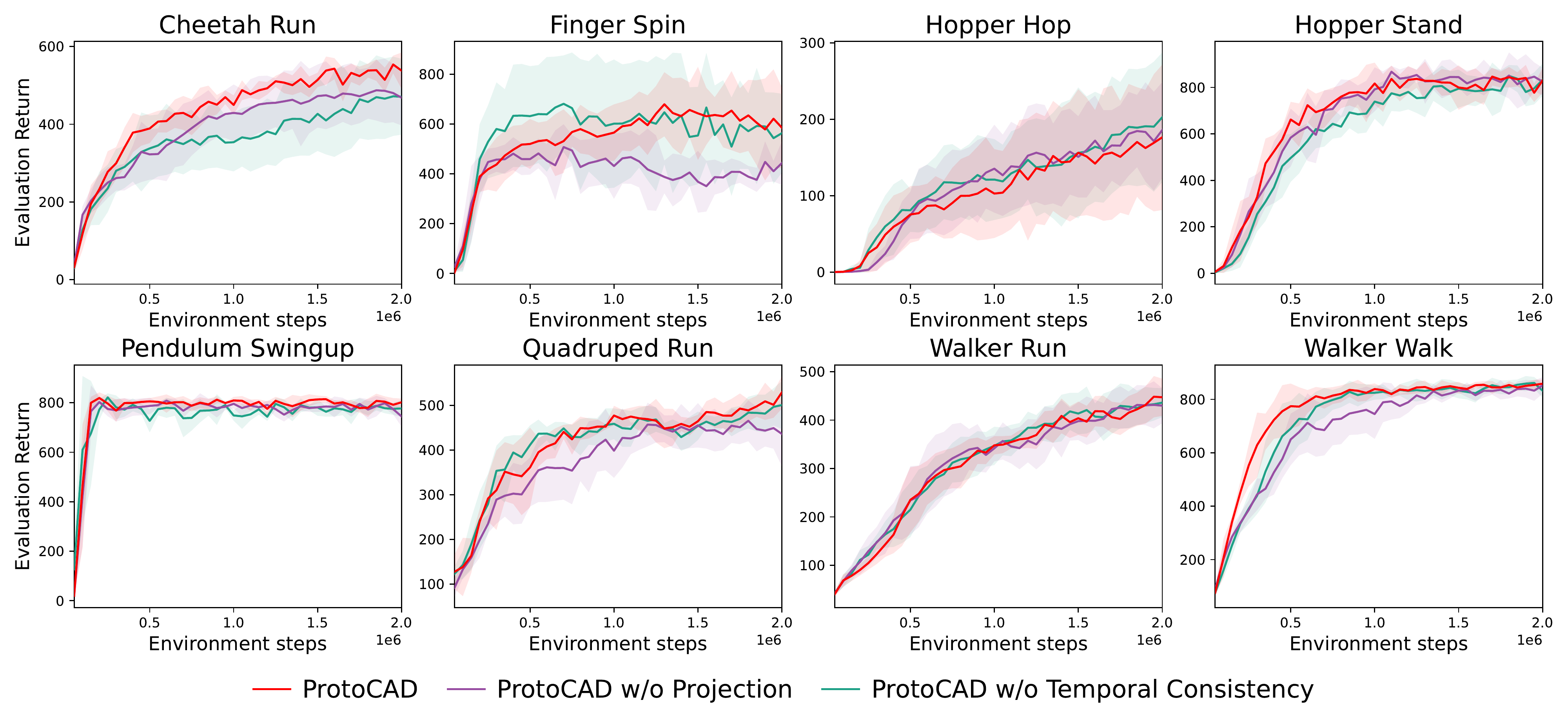}}
\vspace{-10pt}
\caption{Ablations performance of ProtoCAD. }
\label{abla_curves}
\end{figure}

We investigate the contribution of each part of ProtoCAD by removing the individual component from it. We compare the performance of ProtoCAD, ProtoCAD w/o Projection, and  ProtoCAD w/o Temporal Consistency. The results of the ablations are shown in Figure~\ref{abla_curves}. 
\label{app_ablation}

\subsubsection{Additional context-based method comparison}
\label{app_byol}
\begin{figure}[!h]
\centerline{\includegraphics[width=1.0\textwidth]{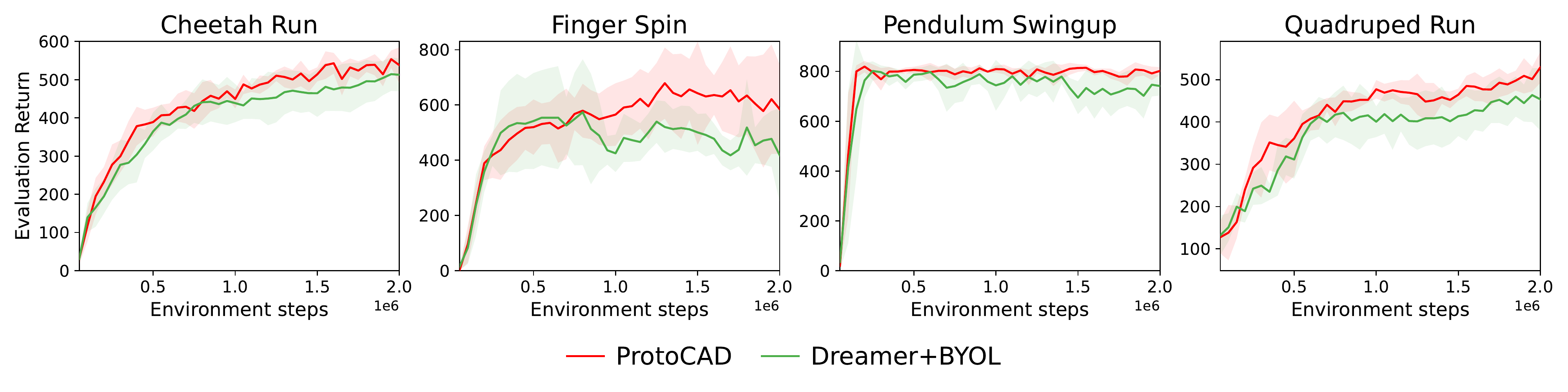}}
\caption{Performance comparison of ProtoCAD and Dreamer+BYOL. The mean and standard deviation are computed from 5 seeds.}
\label{byol_curves}
\end{figure}

We further implement an additional context-aware model-based agent by combining Dreamer with BYOL~\citep{grill2020bootstrap}. 
As shown in Figure~\ref{byol_curves}, ProtoCAD also outperforms Dreamer+BYOL, indicating that the extraction of context features from the latent space by self-supervision alone is not effective enough. This is further evidence of the advantage of prototypical approach in dynamics generalization tasks. 

~\\


\subsubsection{State-based MuJoCo environments evaluation}

The original CaDM, TMCL, and RIA utilize planning to obtain actions and assume a known reward function on state transitions during the planning process, which makes them difficult to apply directly to image input tasks. We replace RSSM in Dreamer with CaDM, implementing a version of CaDM for image input, and the results do not work well enough to learn valid policies. However, we can compare our approach to these methods in the state-based MuJoCo environments with the following results (where the results for CaDM, TMCL and RIA are obtained directly in the RIA paper). 

\begin{table}[htb]
  \renewcommand\arraystretch{1.1}
  \centering
  \footnotesize
  \vspace{-10pt}
  \caption{Performance comparison on state-based MuJoCo environments with unseen dynamics. Here we use $\pm$ to denote the standard deviation. We report the average rewards of ProtoCAD over 5 seeds (Dreamer is 3).}
  \label{state}
  \vspace{10pt}
\begin{tabular}{l|c|c|c|c|c}
\hline
                 & CaDM            & TMCL                    & RIA       & Dreamer             & ProtoCAD (ours) \\
\hline
Pendulum      & $-713.95_{\pm21.1}$  & $-691.2_{\pm93.4}$           & $-587.5_{\pm64.4}$ & $-575.9_{\pm56.6}$ & $\boldsymbol{-525.9}_{\pm61.6}$          \\
Ant      & $1660_{\pm57.8}$ & $2994.9_{\pm243.8}$ & $3297.9_{\pm159.7}$  & $4636.3_{\pm412.8}$ &$\boldsymbol{5309.4}_{\pm537.7}$ \\
Hopper       & $845.2_{\pm20.41}$  & $999.35_{\pm22.8}$           & $1057.4_{\pm37.2}$ & $2107.3_{\pm36.1}$ & $\boldsymbol{2197.8}_{\pm83.44}$           \\
HalfCheetah     & $5876.6_{\pm799.0}$  & $9039.6_{\pm1065}$ & $\boldsymbol{10859.2}_{\pm465.1}$ & $4701.0_{\pm796.2}$ & $5409.8_{\pm594.1}$          \\
C\_HalfCheetah & $3656.4_{\pm856.2}$  & $3998.8_{\pm856.2}$           & $\boldsymbol{4819.3}_{\pm409.3}$  & $4593.0_{\pm654.2}$ & $4125.4_{\pm957.9}$         \\
Slim\_Humanoid    & $859.1_{\pm24.01}$  & $2098.7_{\pm109.1}$           & $2432.6_{\pm465.1}$ & $14735_{\pm5842.9}$ & $\boldsymbol{16731.6}_{\pm9595.8}$         \\
\hline
\end{tabular}
\end{table}

For state inputs, we use random amplitude scaling \citep{laskin2020reinforcement} for augmentation. The results show that our method substantially outperforms the SOTA method RIA in 4/6 environments. It is worth noting that Dreamer also performs well, showing the ability of RSSM itself to handle in dynamics generalization problems. This also shows that Dreamer is a strong baseline. 

\subsubsection{Performance on specific parameter settings}

\begin{figure}[ht]
\centering
\subfigure[Dreamer]{
\begin{minipage}[b]{0.31\textwidth}
\centering
\includegraphics[width=1.0\textwidth]{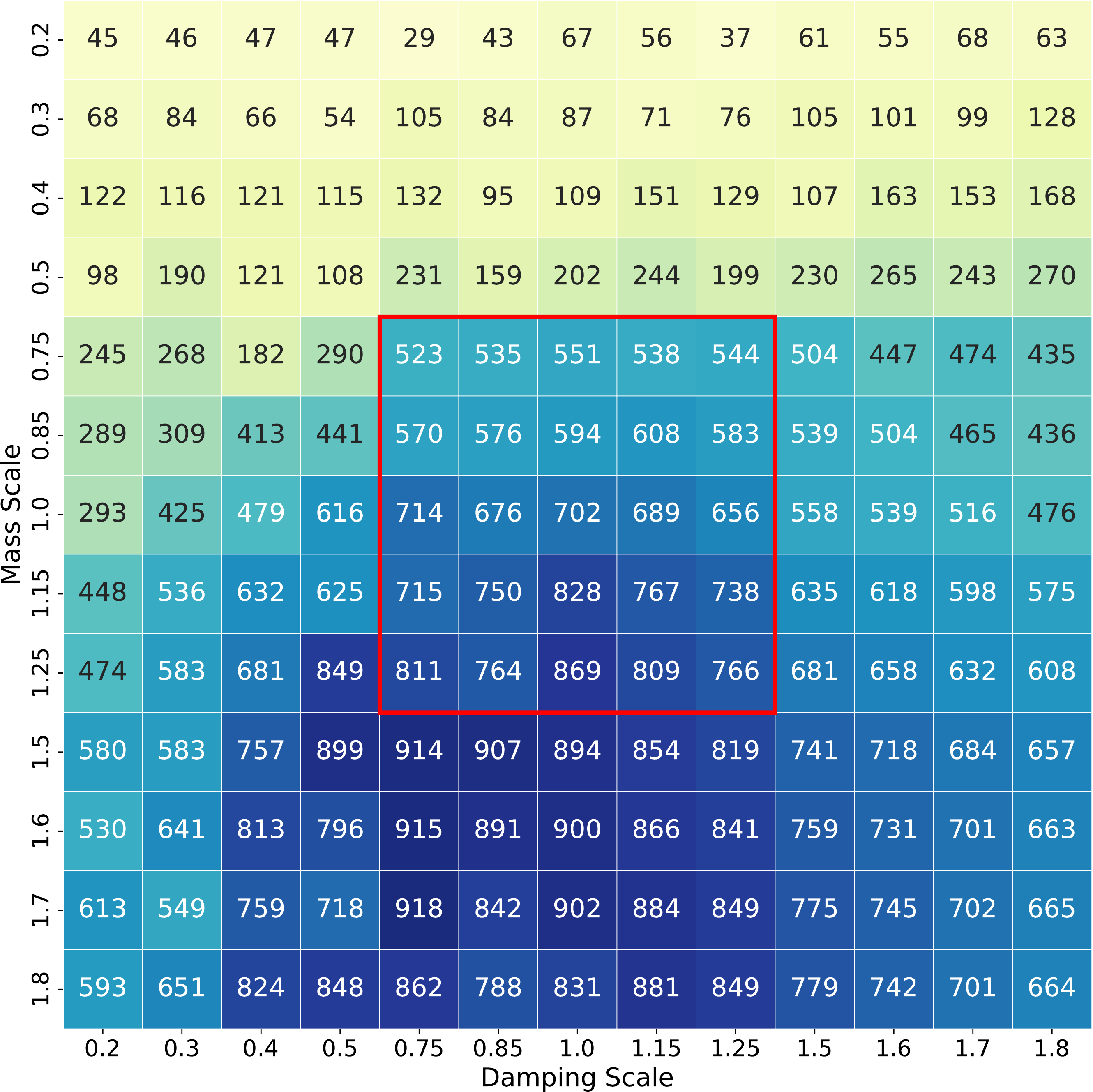}
\end{minipage}
}
\subfigure[DreamerPro]{
\begin{minipage}[b]{0.31\textwidth}
\centering
\includegraphics[width=1.0\textwidth]{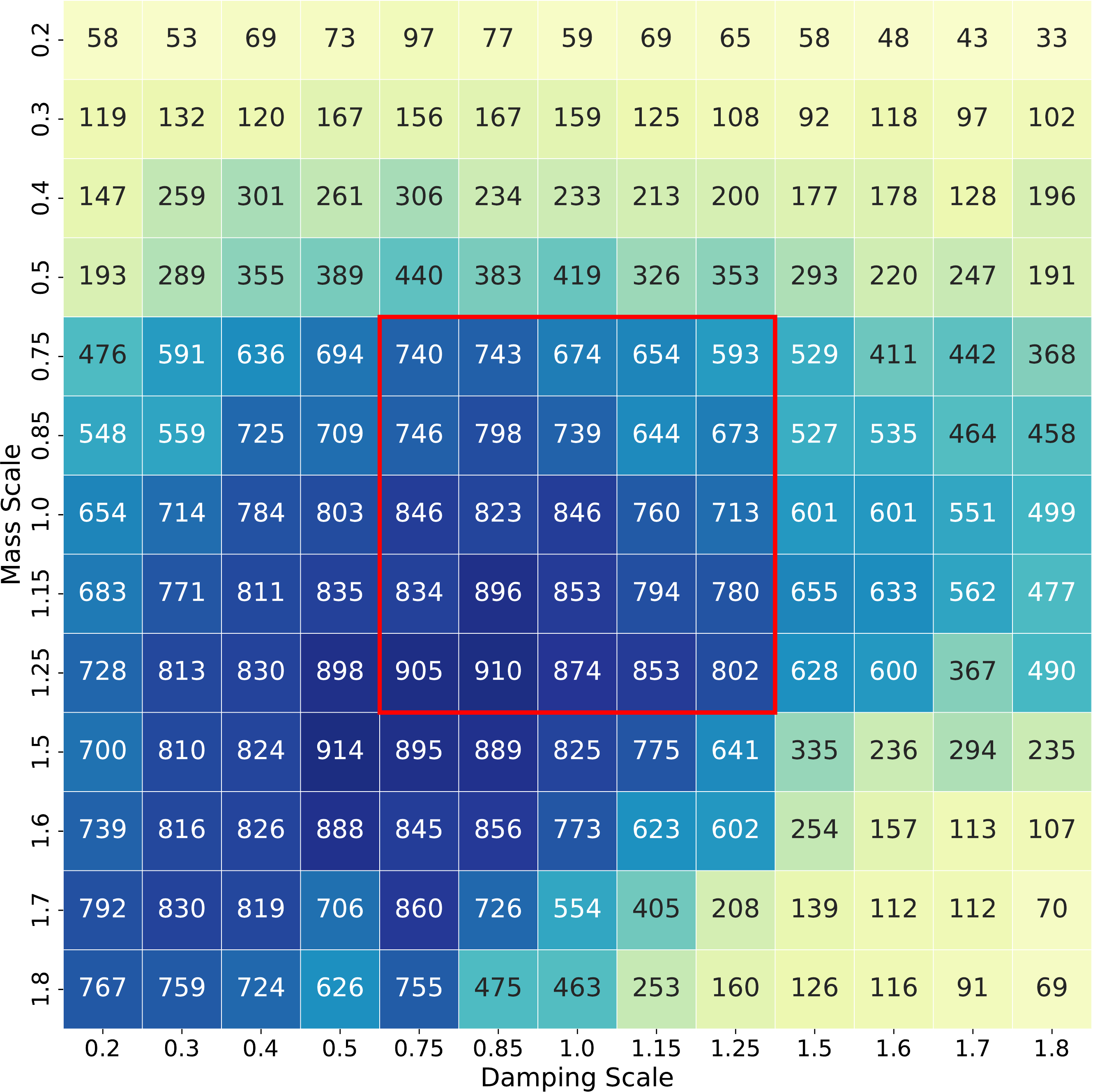}
\end{minipage}
}
\subfigure[ProtoCAD (ours)]{
\begin{minipage}[b]{0.31\textwidth}
\centering
\includegraphics[width=1.0\textwidth]{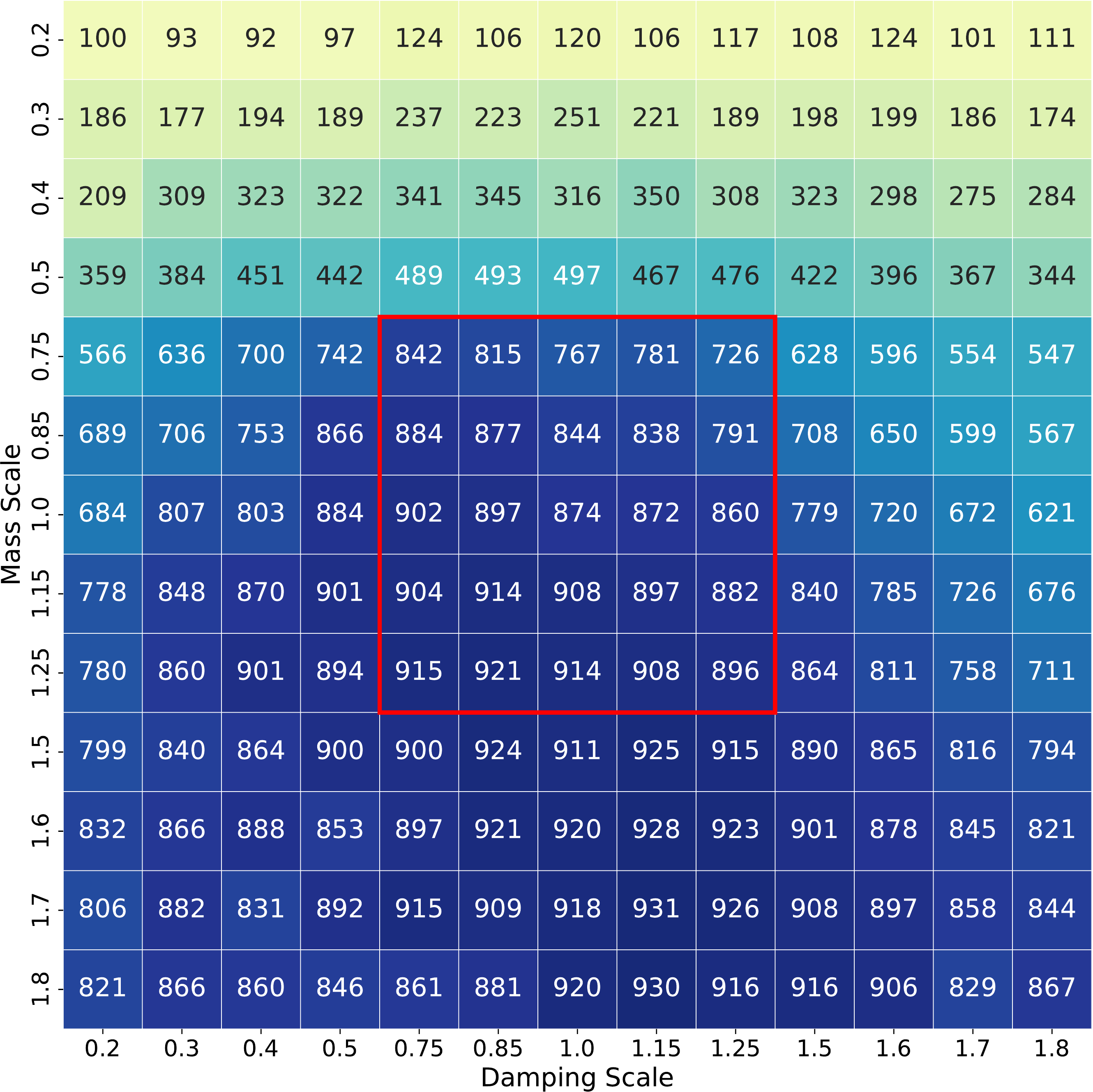}
\end{minipage}
}
\centering
\vspace{-10pt}
\caption{Performance on the task of Cheetah Run with varying dynamics.}
\vspace{-10pt}
\label{test}
\end{figure}

Similar to literature \citep{ball2021augmented}, we test different dynamics environments constructed for specific parameter settings separately in the Cheetah Run task. Results are given in Figure~\ref{test}. Note that the red box shows the parameter settings during training. From the results, we can see that our method has better generalization to different parameter settings. 

\subsubsection{TSNE visualization}

\begin{figure}[ht]
\centerline{\includegraphics[width=0.6\textwidth]{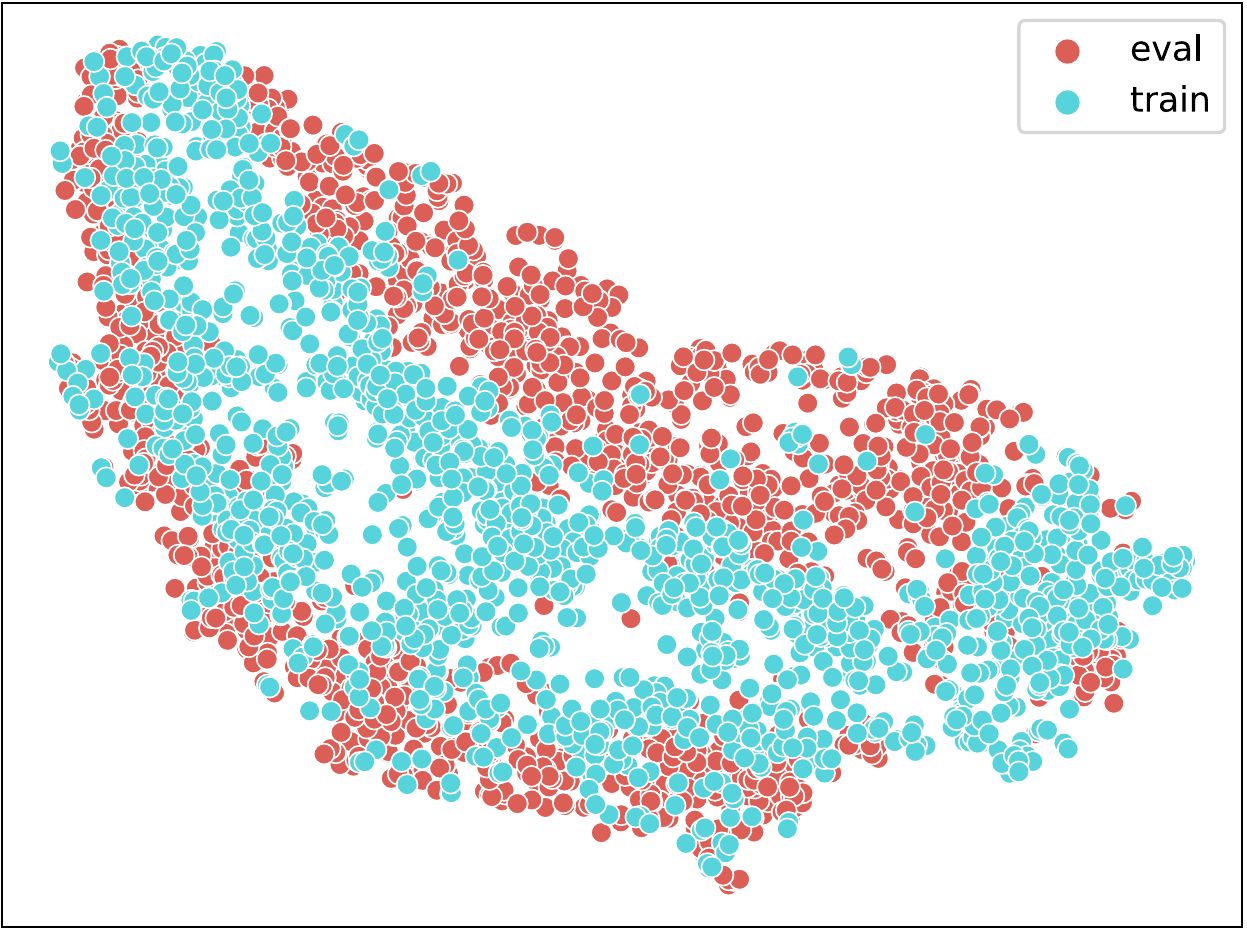}}
\vspace{-10pt}
\caption{TSNE result of the learned feature for training and testing. After the training is completed, we extract features from the training and test data respectively, which is the concatenate of the projection output $u$ and the aggregated prototypes based on the prediction probabilities $w$. We use TSNE to perform dimensionality reduction on this feature, and the visualization shows that the representation is in the vicinity of the training representation when tested in unseen environments, indicating that the context representation can be generalized.
}
\vspace{-10pt}
\label{tsne_train}
\end{figure}

\begin{figure}[ht]
\centering
\subfigure[RSSM State]{
\begin{minipage}[b]{0.45\textwidth}
\centering
\includegraphics[width=0.95\textwidth]{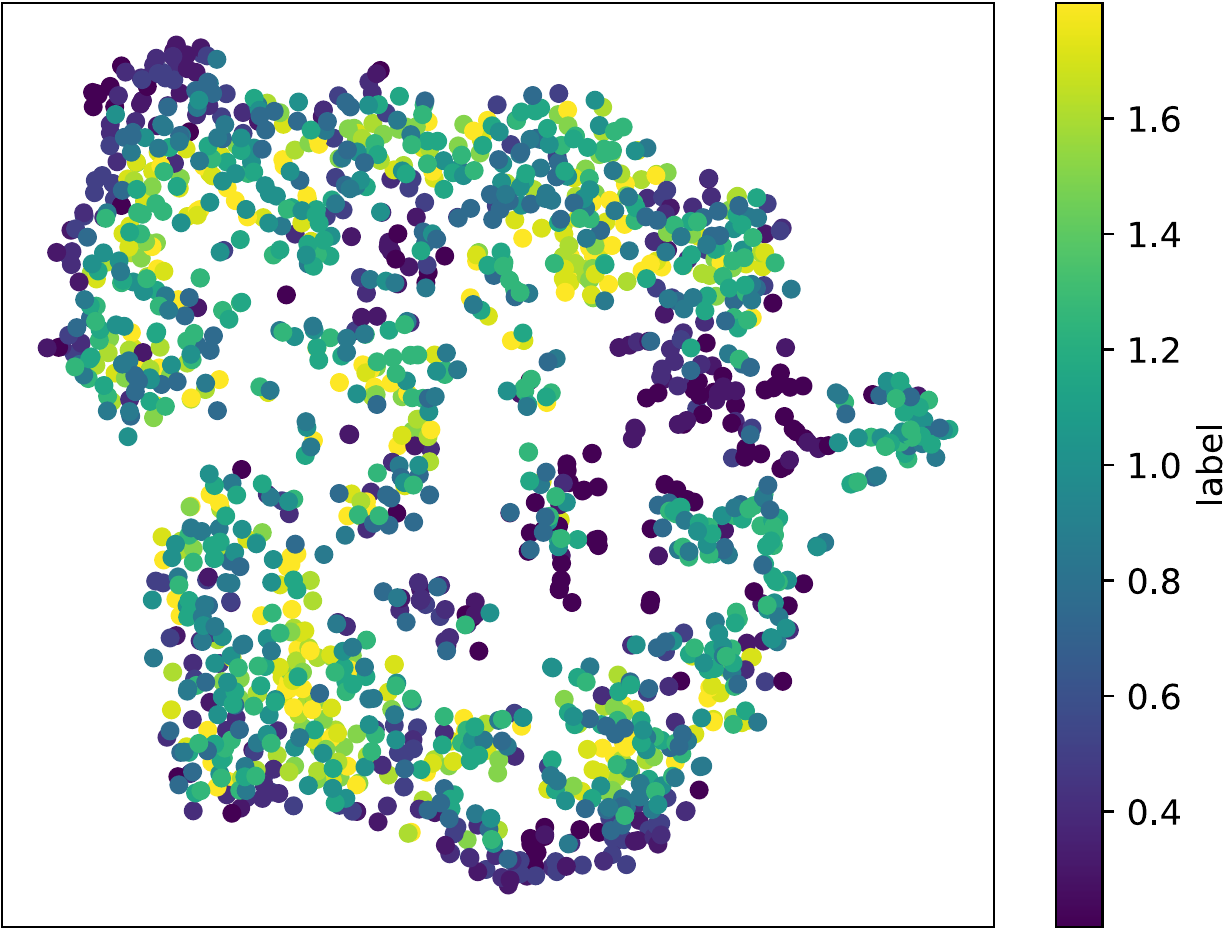}
\end{minipage}
}
\subfigure[Learned Feature]{
\begin{minipage}[b]{0.45\textwidth}
\centering
\includegraphics[width=0.95\textwidth]{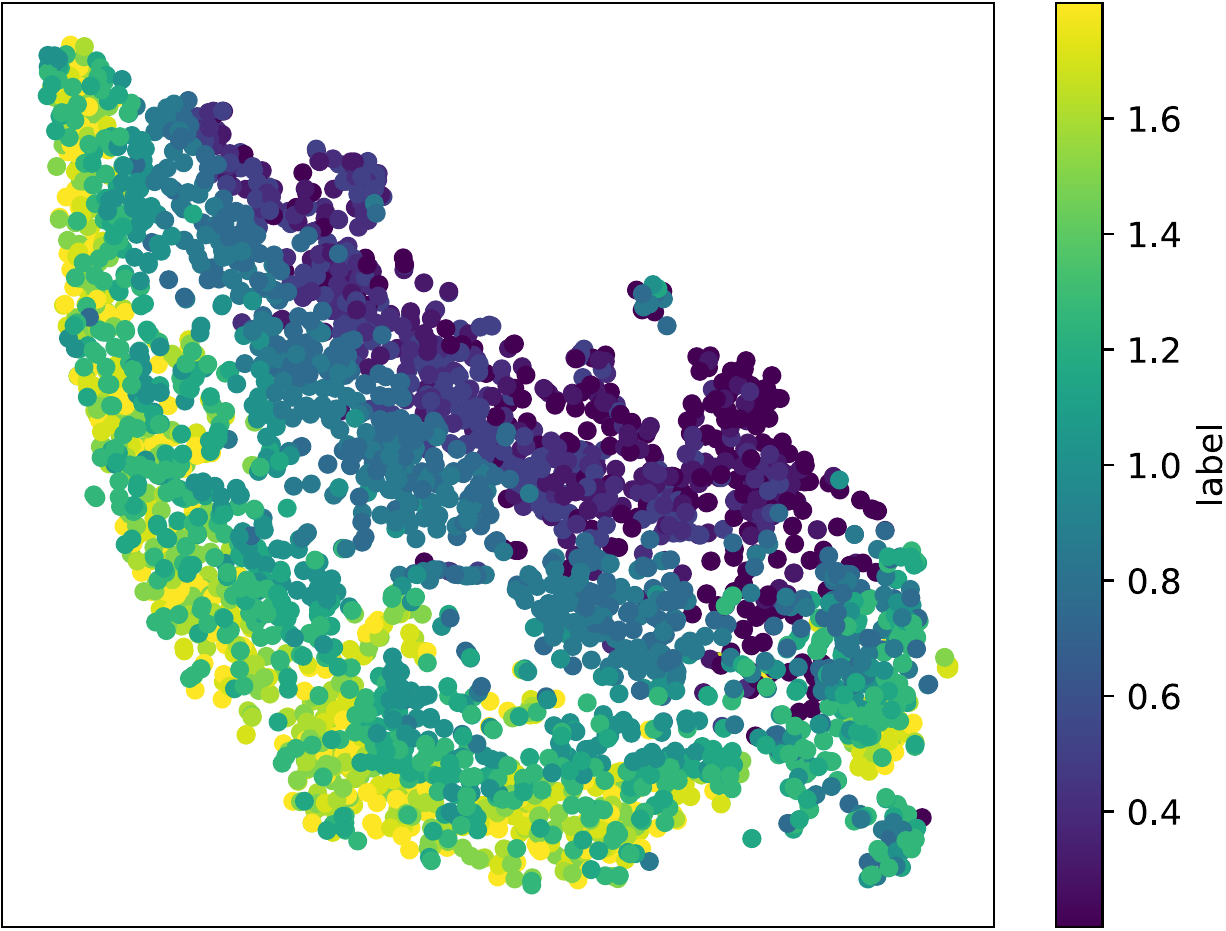}
\end{minipage}
}
\centering
\vspace{-10pt}
\caption{TSNE results of RSSM state and learned feature for different parameter settings. Compared with the original RSSM state, the context representation has a significant clustering result for different parameter settings. }
\vspace{-10pt}
\label{tsne}
\end{figure}

Our motivation is to extract the environment context information from the state trajectories produced by RSSM to assist in policy learning. Sinkhorn-Knopp can cluster trajectory data of batch and use prototypes to fit the different situations encountered in the learning process. The learned prototypes can characterize the context information. The results from TSNE (see Figure~\ref{tsne_train} and Figure~\ref{tsne}) show that the learned context representation has some differentiated characterization results for different parameter settings. Also, the representation can be generalized to new parameter settings when testing under unseen environments. The learned representation is used as part of the input to the policy network and the value network so that the policy has some generalization ability in new environments as well. 

\end{document}

%% file: math_commands.tex
\usepackage{amsmath,amsfonts,bm}

\def\eqref#1{equation~\ref{#1}}

\def\1{\bm{1}}

\DeclareMathAlphabet{\mathsfit}{\encodingdefault}{\sfdefault}{m}{sl}
\SetMathAlphabet{\mathsfit}{bold}{\encodingdefault}{\sfdefault}{bx}{n}

